\documentclass[journal]{IEEEtran}
\ifCLASSINFOpdf
\else
\fi

\usepackage{amsmath}
\usepackage{amsfonts}
\usepackage{graphicx, subfigure}
\usepackage{algorithm, algorithmic}
\usepackage{color}
\usepackage{multirow}

\newcommand{\D}[2]{\frac{\partial #1}{\partial #2}}
\newcommand{\DD}[2]{\frac{\partial^2 #1}{\partial {#2}^2}}
\newcommand{\red}[1]{\textcolor{red}{#1}}

\graphicspath{{figs/}}

\begin{document}

\title{GBDT-MO: Gradient Boosted Decision Trees \\ for Multiple Outputs}

\author{Zhendong~Zhang
        and Cheolkon~Jung,~\IEEEmembership{Member,~IEEE}
\thanks{This work was supported by the National Natural Science Foundation of China (No. 61872280) and the International S\&T Cooperation Program of China (No. 2014DFG12780).}
\thanks{Z. Zhang and C. Jung (corresponding author) are with the School of Electronic Engineering, Xidian University, Xi'an 710071, China e-mail: zhd.zhang.ai@gmail.com, zhengzk@xidian.edu.cn}
}
\markboth{Under Review}%
{Zhang and Jung: Gradient Boosted Decision Trees for Multiple Outputs}

\maketitle

\begin{abstract}
Gradient boosted decision trees (GBDTs) are widely used in machine learning, and the output of current GBDT implementations is a single variable. When there are multiple outputs, GBDT constructs multiple trees corresponding to the output variables. The correlations between variables are ignored by such a strategy causing redundancy of the learned tree structures. In this paper, we propose a general method to learn GBDT for multiple outputs, called GBDT-MO. Each leaf of GBDT-MO constructs predictions of all variables or a subset of automatically selected variables. This is achieved by considering the summation of objective gains over all output variables. Moreover, we extend histogram approximation into multiple output case to speed up the training process. Various experiments on synthetic and real-world datasets verify that GBDT-MO achieves outstanding performance in terms of both accuracy and training speed. Our codes are available on-line.

\end{abstract}


%
\IEEEpeerreviewmaketitle

\section{Introduction}
\IEEEPARstart{M}{achine} learning and data-driven approaches have achieved great success in recent years. Gradient boosted decision tree (GBDT) \cite{friedman2001greedy} \cite{friedman2002stochastic} is a powerful machine learning tool widely used in many applications including multi-class classification \cite{li2010robust}, flocculation process modeling \cite{qi2018data-driven}, learning to rank \cite{burges2010ranknet} and click prediction \cite{richardson2007predicting}. It also produces state-of-the-art results for many data mining competitions such as the Netflix prize \cite{bennett2007netflix}. GBDT uses decision trees as the base learner and sums the predictions of a series of trees. At each step, a new decision tree is trained to fit the residual between ground truth and current prediction. GBDT is popular due to its accuracy, efficiency and interpretability. Many improvements have been proposed after \cite{friedman2001greedy}. XGBoost \cite{chen2016xgboost:} used the second order gradient to guide the boosting process and improve the accuracy. LightGBM \cite{ke2017lightgbm:} aggregated gradient information in histograms and significantly improved the training efficiency. CatBoost \cite{prokhorenkova2018catboost:} proposed a novel strategy to deal with categorical features.

A limitation of current GBDT implementations is that the output of each decision tree is a single variable. This is because each leaf of a decision tree produces a single variable. However, multiple outputs are required for many machine learning problems including but not limited to multi-class classification, multi-label classification \cite{mulan} and multi-output regression \cite{borchani2015a}. Other machine learning methods, such as neural networks \cite{Goodfellow-et-al-2016}, can adapt to any dimension of outputs straightforwardly by changing the number of neurons in the last layer. The flexibility for the output dimension may be one of the reasons why neural networks are popular. However, it is somewhat strange to handle multiple outputs by current GBDT implementations. At each step, they construct multiple decision trees each of which corresponds to an individual output variable, then concatenates the predictions of all trees to obtain multiple outputs. This strategy is used in the most popular open-sourced GBDT libraries: XGBoost \cite{chen2016xgboost:}, LightGBM \cite{ke2017lightgbm:}, and CatBoost \cite{prokhorenkova2018catboost:}.

The major drawback of the abovementioned strategy is that correlations between variables are ignored during the training process because those variables are treated in isolation and they are learned independently. However, correlations more or less exist between output variables. For example, there are correlations between classes for multi-class classification. It is verified in \cite{hinton2015distilling} that such correlations improve the generalization ability of neural networks. Ignoring variable correlations also leads to redundancy of the learned tree structures. Thus, it is necessary to learn GBDT for multiple outputs via better strategies. Up to now, a few works have explored it. Geurts et al. \cite{geurts2007gradient} transformed the multiple output problem into a single output problem by kernelizing the output space. However, this method was not scalable because the space complexity of its kernel matrix was $n^2$ where $n$ is the number of training samples. Si et al. \cite{si2017gradient} proposed GBDT for sparse output (GBDT-sparse). They mainly focused on extreme multi-label classification problems, and the outputs were represented in sparse format. A sparse split finding algorithm was designed for square hinge loss. \cite{geurts2007gradient} and \cite{si2017gradient} worked for specific loss and they did not employ the second order gradient and histogram approximation.

In this paper, we propose a novel and general method to learn GBDT for multiple outputs, which is scalable and efficient, named GBDT-MO. Unlike previous works, we employ the second order gradient and histogram approximation to improve GBDT-MO. The learning mechanism is designed based on them to jointly fit all variables in a single tree. Each leaf of a decision tree constructs multiple outputs at once. This is achieved by maximizing the summation of objective gains over all output variables. Sometimes, only a subset of the output variables is correlated. It is expected that the proposed method automatically selects those variables and constructs predictions for them at a leaf. We achieve this by adding $L_0$ constraint to the objective function. Since the learning mechanism of GBDT-MO enforces the learned trees to capture variable correlations, it plays a role in indirect regularization. Experiments on both synthesis and real-world datasets show that GBDT-MO achieves better generalization ability than the standard GBDT. Moreover, GBDT-MO achieves a fast training speed, especially when the number of outputs is large.

\begin{figure*} [t]
	\centering
	\subfigure[Round 1, variable 1]{
		\includegraphics[width=0.33\textwidth]{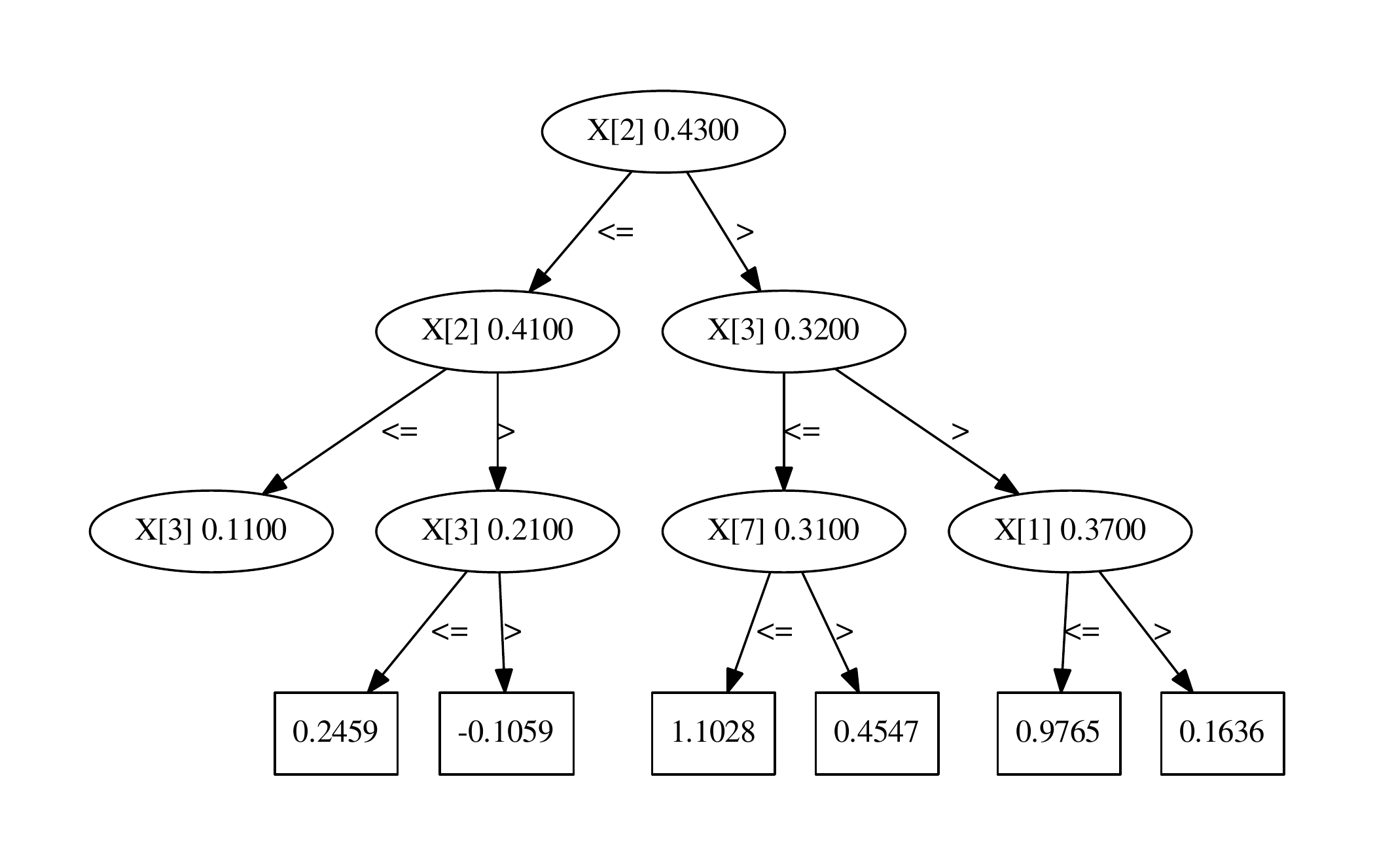}
	}
	\subfigure[Round 1, variable 2]{
		\includegraphics[width=0.33\textwidth]{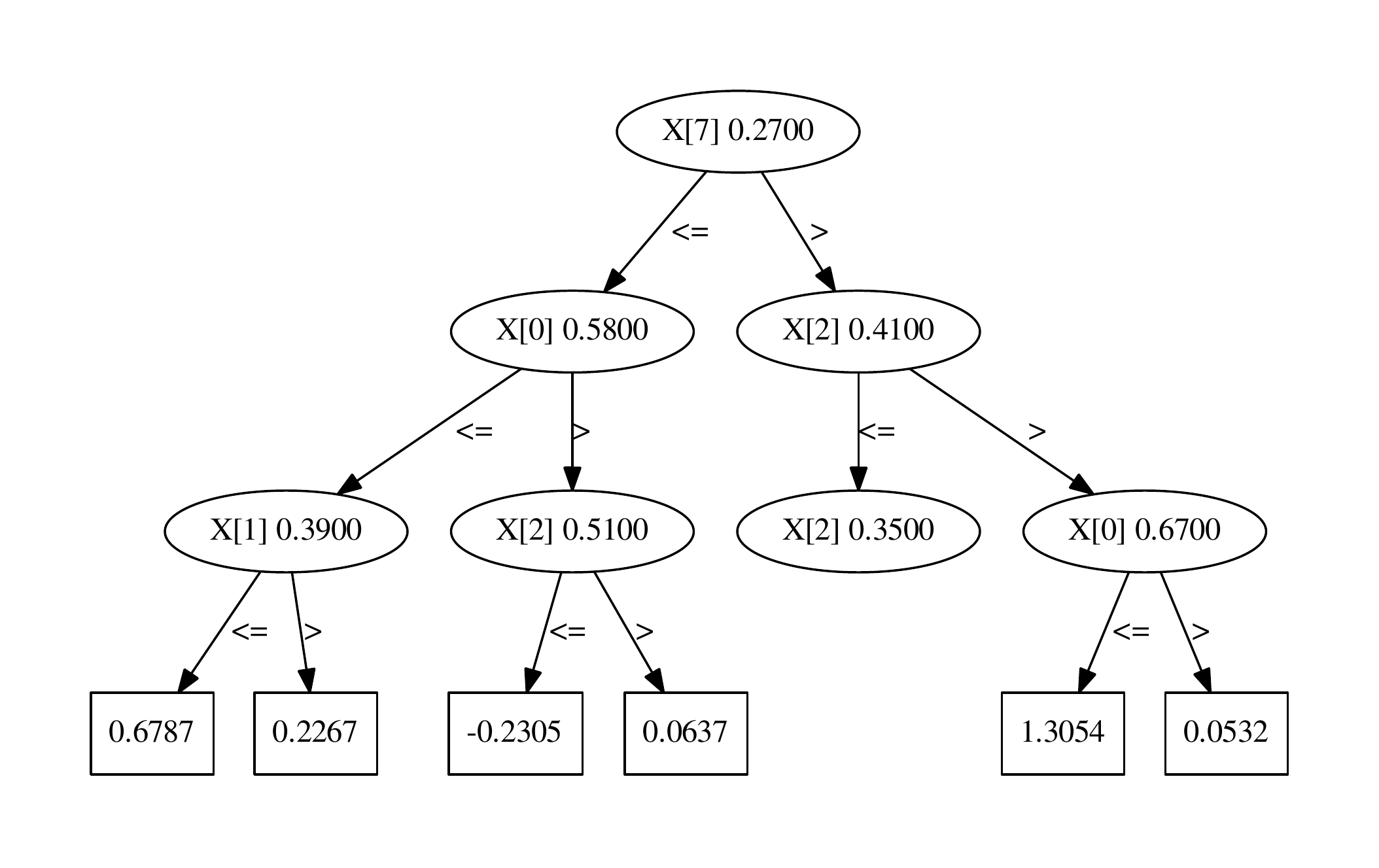}
	}
	\subfigure[Round 1]{
		\includegraphics[width=0.25\textwidth]{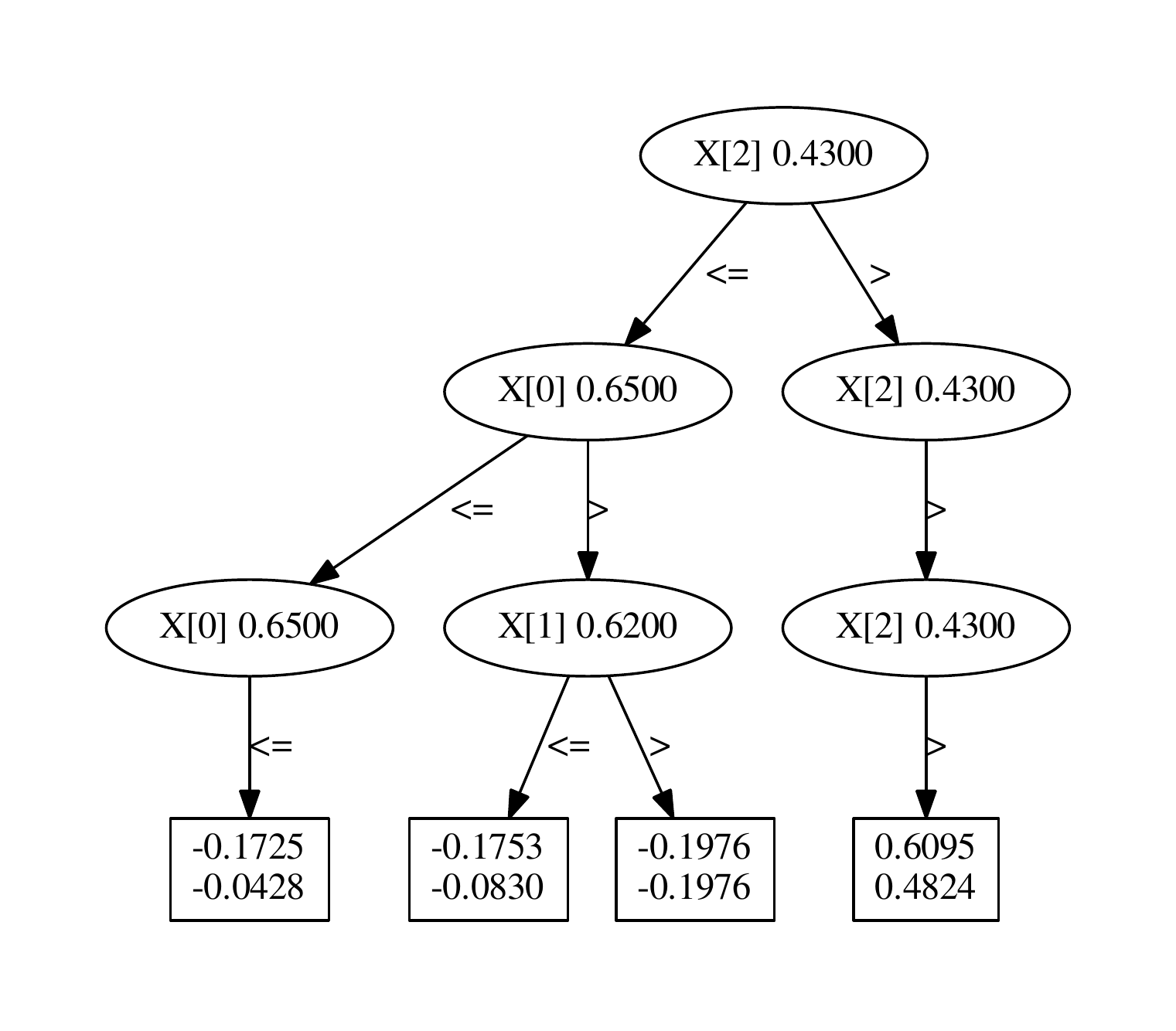}
	}
	\subfigure[Round 2, variable 1]{
		\includegraphics[width=0.31\textwidth]{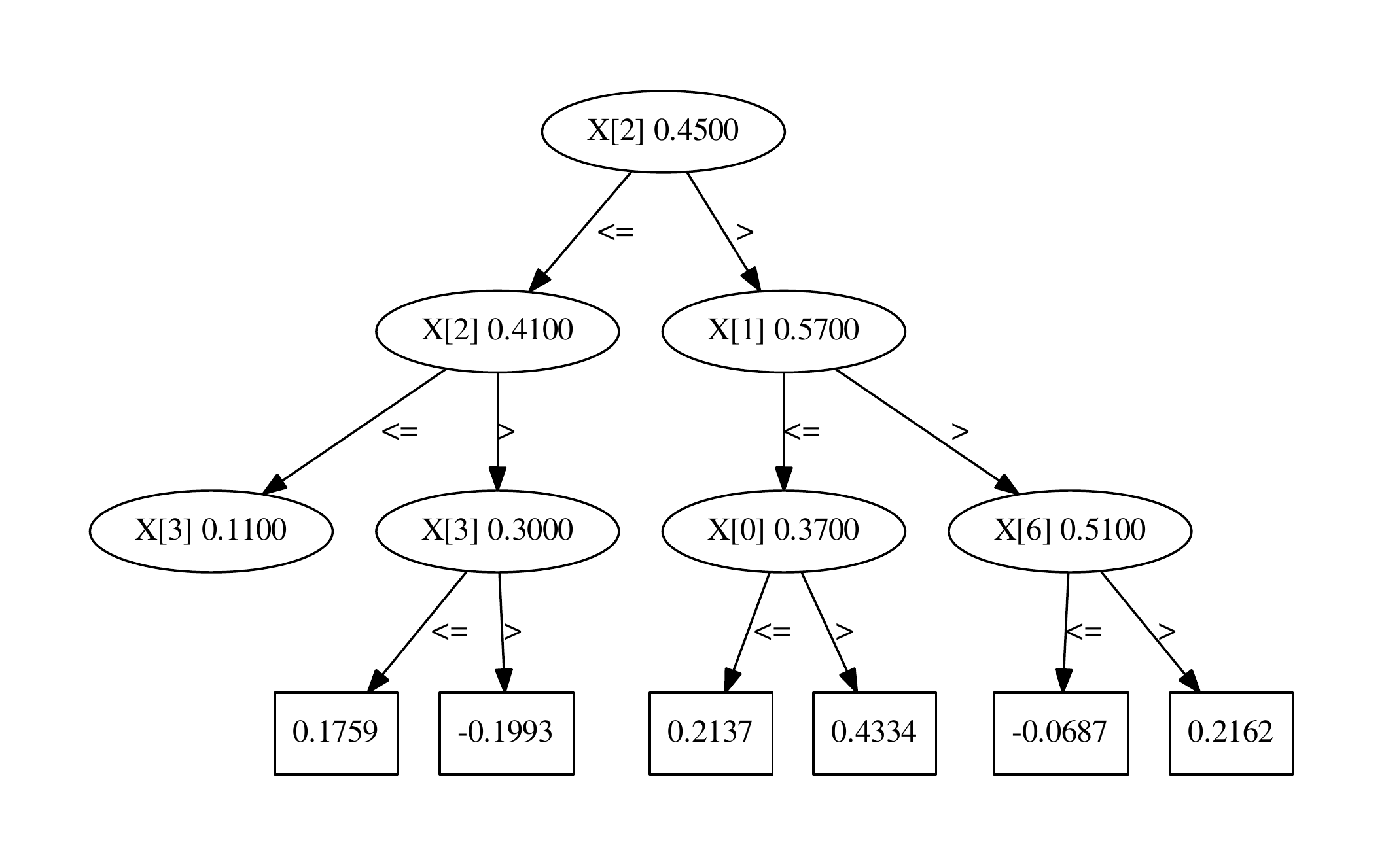}
	}
	\subfigure[Round 2, variable 2]{
		\includegraphics[width=0.31\textwidth]{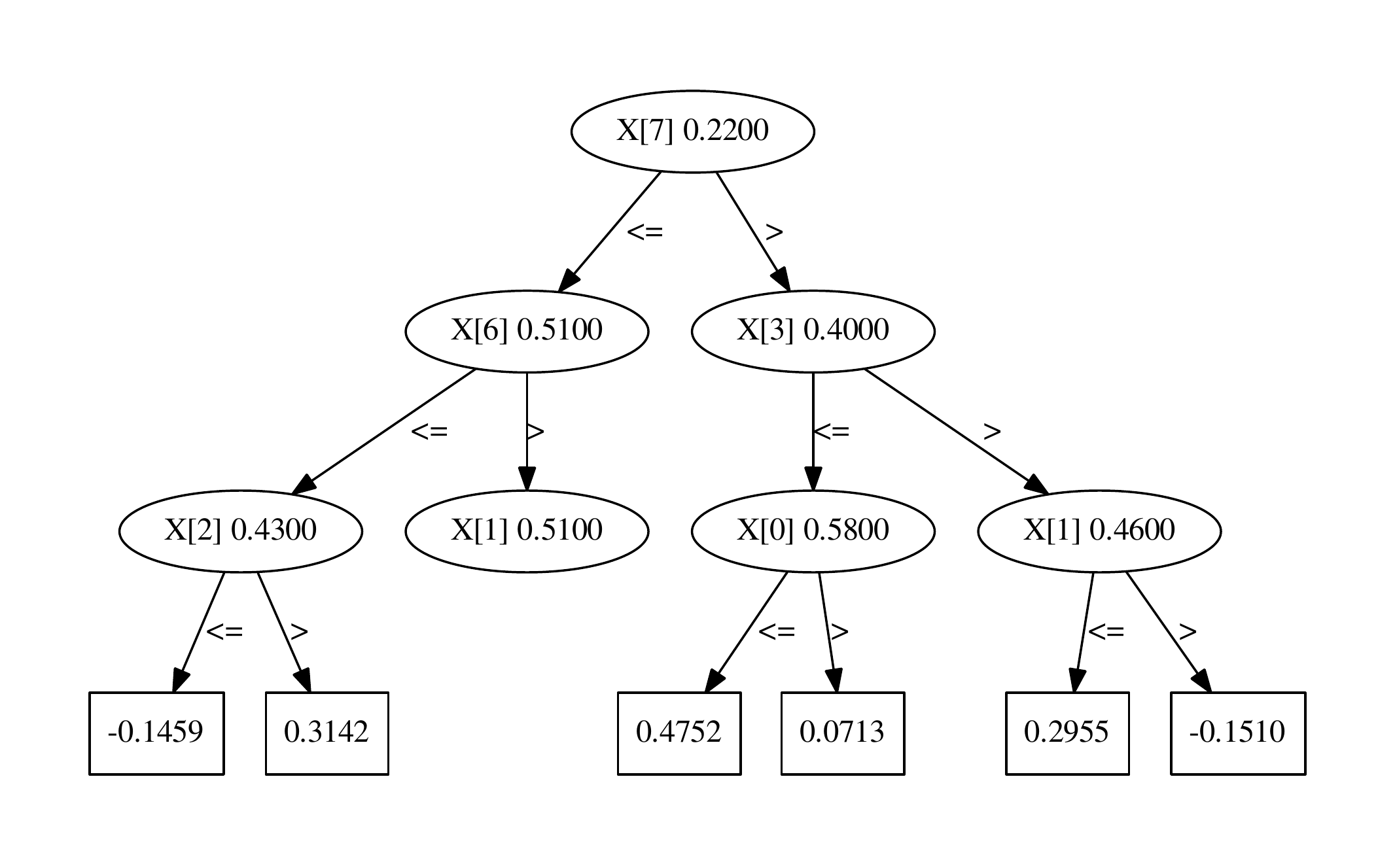}
	}
	\subfigure[Round 2]{
		\includegraphics[width=0.31\textwidth]{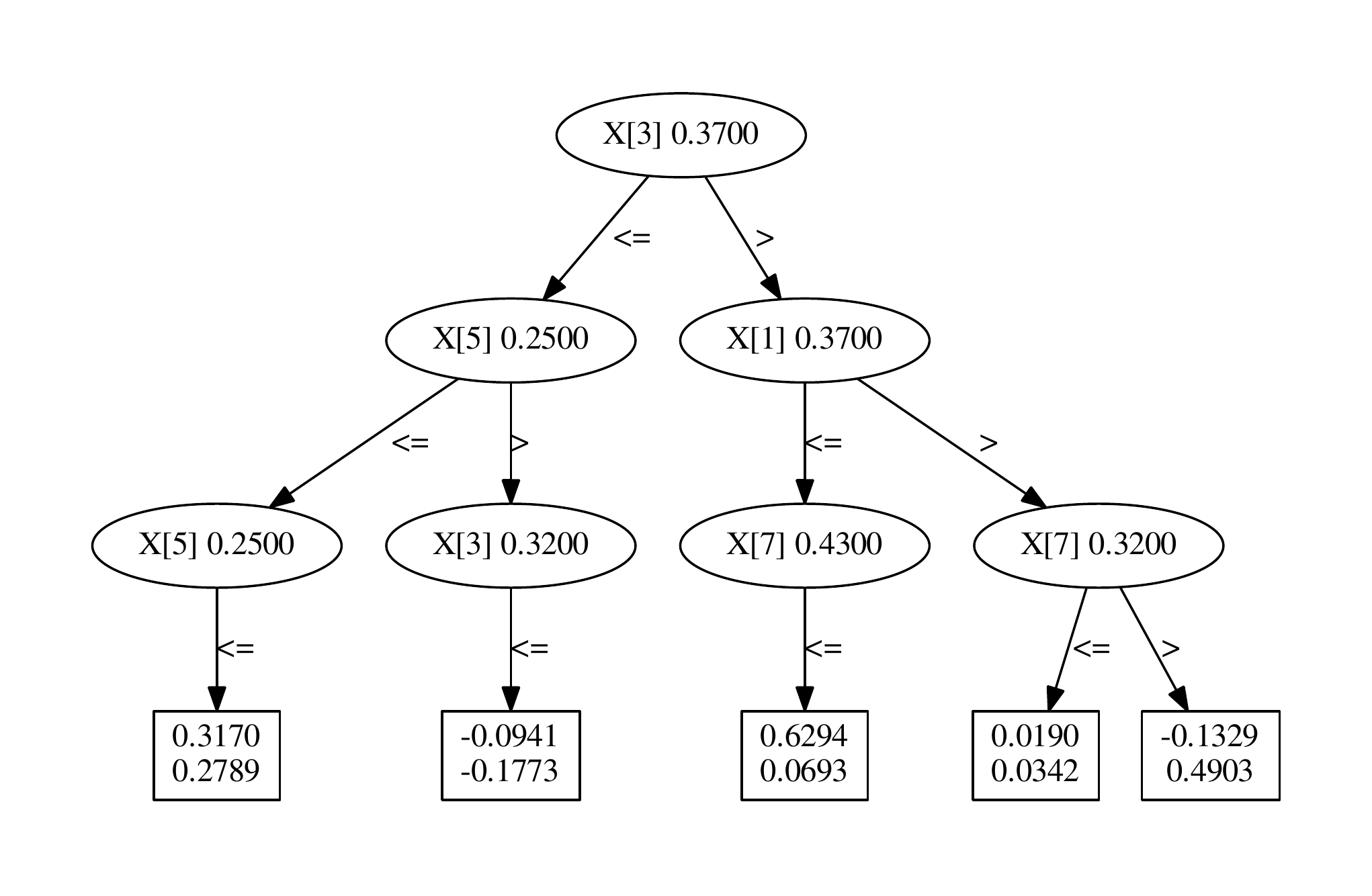}
	}
	\caption{Illustration of the learned trees of GBDT-SO and GBDT-MO on \textbf{Yeast} with 10 output variables. The first and second columns show GBDT-SO for the first and second output variables, respectively. The last column shows GBDT-MO for corresponding variables. We limit tree depth to 3 for illustration.}
\end{figure*}

Compared with existing methods, main contributions of this paper are as follows:
\begin{itemize}
	\item We formulate the problem of learning multiple outputs for GBDT, and propose a split finding algorithm by deriving a general approximate objective for this problem.
	\item To learn a subset of outputs, we add a sparse constraint to the objective. Based on it, we develop two sparse split finding algorithms.
	\item We extend histogram approximation \cite{tyree2011parallel} into multiple output case to speed up the training process.
\end{itemize}

The rest of this paper is organized as follows. First, we review GBDT for single output and introduce basic definitions in Section \ref{sec_single}. Then, we describe the details of GBDT-MO in Section \ref{sec_multi}. We address related work in Section \ref{sec_related}. Finally, we perform experiments and conclude in Sections \ref{sec_exp} and \ref{sec_conc}, respectively.

\section{GBDT for Single Output}
\label{sec_single}

In this section, we review GBDT for single output. First, we describe the work flow of GBDT. Then show how to derive the objective of GBDT based on the second order Taylor expansion of the loss, which is used in XGBoost. The objective to multiple variable cases will be generalized in Section \ref{sec_multi}. Finally, we explain the split finding algorithms which exactly or approximately minimize the objective.

\subsection{Work Flow}
Denote $\mathcal{D} = \{ (\mathbf{x}_i, y_i)_{i=1}^n\}$ as a dataset with $n$ samples, where $\mathbf{x} \in \mathbb{R}^m$ is an $m$ dimension input. Denote $f : \mathbb{R}^m \rightarrow \mathbb{R}$ as the function of a decision tree which maps $\mathbf{x}$ into a scalar. Since GBDT integrates $t$ decision trees in an additive manner, the prediction of GBDT is $\hat{y}_i = \sum_{k=1}^{t} f_k(\mathbf{x}_i)$, where $f_k$ is the function of $k$-th decision tree. GBDT aims at constructing a series of trees given datasets. It first calculates gradient based on current prediction at each boosting round, and then constructs a new tree guided by gradient. Finally, it updates the prediction using the new tree. The most important part of GBDT is to construct trees based on gradient.

\subsection{Objective}
 Based on the construction mechanism of decision trees, $f$ can be further expressed as follows:
\begin{equation}
f(\mathbf{x}) = \mathbf{w}_{q(\mathbf{x})}, \quad q : \mathbb{R}^m \rightarrow [1, L], \quad \mathbf{w} \in \mathbb{R}^L
\end{equation}
where $L$ is the number of leaves of a decision tree, $q$ is a function which selects a leaf given $\mathbf{x}$ and $\mathbf{w}_i$ is the value of $i$-th leaf. That is, once a decision tree is constructed, it first maps input into a leaf, then returns the value of that leaf.

Now, we consider the objective of $(t+1)$-th decision tree given prediction $\hat{y}$ of the first $t$ tree.
\begin{equation}
\label{eq_obj_f}
\sum_{i=1}^{n} l (\hat{y}_i + f(\mathbf{x}_i), y_i) + \lambda \mathcal{R}(f)
\end{equation}
where the first term is fidelity term, $\mathcal{R}$ is regularization term of $f$. In this work, $\lambda$ is a positive number to control the trade-off between fidelity term and regularization term. We suppose $l$ is a second order differentiable loss. Based on the space of $f$, i.e. a constant value for each leaf, the fidelity term of \eqref{eq_obj_f} is separable w.r.t. each leaf. Then, \eqref{eq_obj_f} is rewritten as follows:
\begin{equation}
\label{eq_obj}
\sum_{j=1}^{L} \left\{ \sum_{i \in leaf_j} l (\hat{y}_i + \mathbf{w}_j, y_i) \right\} + \lambda \mathcal{R}(\mathbf{w})
\end{equation}
Although there are many choices of $\mathcal{R}$, we set $\mathcal{R}(\mathbf{w}) = \frac{1}{2}\lVert \mathbf{w} \rVert_2^2$, which is commonly used. Because \eqref{eq_obj} is \emph{separable} w.r.t. each leaf, we only consider the objective of a single leaf as follows:
\begin{equation}
\label{eq_obj_leaf}
\mathcal{L} = \sum_i l(\hat{y}_i + w, y_i) + \frac{\lambda}{2}w^2
\end{equation}
where $w$ is the value of a leaf and $i$ is enumerated over the samples belonging to that leaf. $l (\hat{y}_i + w, y_i)$ can be approximated by the second order Taylor expansion of $l (\hat{y}_i, y_i)$. Then, we have
\begin{equation}
\label{eq_obj_taylor}
\mathcal{L} = \sum_i \left\{l (\hat{y}_i, y_i) + g_iw + \frac{1}{2}h_iw^2 \right\} + \frac{\lambda}{2}w^2
\end{equation}
where $g_i$ and $h_i$ are the first and second order derivatives of $l(\hat{y}_i, y)$ w.r.t. $\hat{y}_i$. By setting $\D{\mathcal{L}}{w}$ to $0$, we obtain the optimal value of $w$ as follows:
\begin{equation}
\label{eq_opt_w}
w^* = - \frac{\sum_i g_i}{\sum_i h_i + \lambda}
\end{equation}
Substituting \eqref{eq_opt_w} into \eqref{eq_obj_taylor}, we get the optimal objective as follows:
\begin{equation}
\label{eq_opt_l}
\mathcal{L}^* = -\frac{1}{2} \frac{\left( \sum_i g_i \right)^2}{\sum_i h_i + \lambda}
\end{equation}
We ignore $l(\hat{y}, y)$ since it is a constant term given $\hat{y}$.

\subsection{Split Finding}

One of the most important problems in decision tree learning is to find the best split given a set of samples. Specifically, samples are divided into left and right parts based on the following rule:

\begin{equation}
\mathbf{x}_i \in \left\{
\begin{array}{cc}
left, \qquad & \mathbf{x}_{ij} \le T \\
right, \qquad & \mathbf{x}_{ij} > T
\end{array}
\right.
\end{equation}
where $\mathbf{x}_{ij}$ is $j$-th element of $\mathbf{x}_i$ and $T$ is the threshold. The goal of split finding algorithms is to find the best column $j$ and threshold $T$ such that the gain between the optimal objectives before split and after split is maximized. The optimal objective after split is defined as the sum of the optimal objectives of left and right parts.
\begin{equation}
gain = \mathcal{L}^* - (\mathcal{L}^*_{left} + \mathcal{L}^*_{right})
\end{equation}
where maximizing $gain$ is equivalent of minimizing $\mathcal{L}^*_{left} + \mathcal{L}^*_{right}$ because $\mathcal{L}^*$ is fixed for a given set of samples. $gain$ is used to determine whether a tree is grown. If $gain$ is smaller than a threshold, we stop the growth to avoid over-fitting.

Exact and approximate split finding algorithms have been developed. The exact algorithm enumerates over all possible splits on all columns. For efficiency, it first sorts the samples according to the values of each column and then visit the samples in the sorted order to accumulate $g$ and $h$ which are used to compute the optimal objective. The exact split finding algorithm is accurate. However, when the number of samples is large, it is time-consuming to enumerate over all of the possible splits. Approximate algorithms are necessary as the number of samples increases. The key idea of approximate algorithms is that they divide samples into buckets and enumerate over these buckets instead of individual samples. The gradient statistics are accumulated within each bucket. The complexity of the enumeration process for split is independent of the number of samples. In literature, there are two strategies for bucketing: quantile-based and histogram-based. Since the latter is significantly faster than the former \cite{ke2017lightgbm:}, we focus on the latter in this work. For histogram-based bucketing, a bucket is called a bin. Samples are divided into $b$ bins by $b$ adjacent intervals: $(s_0, s_1, s_2, \dots, s_b)$ where $s_0$ and $s_b$ are usually set to $-\infty$ and $+\infty$ respectively. These intervals are constructed based on the distribution of $j$th input column of the whole dataset. Once constructed, they are keeping unchanged. Given a sample $\mathbf{x}_i$, it belongs to $k$th bin if and only if $s_{k-1} < \mathbf{x}_{ij} <= s_k$. The bin value of $\mathbf{x}_{ij}$ is obtained by binary-search with complexity $\mathcal{O}(\log b)$. Given bin values, the histogram of $j$-th input column is constructed by a single fast scanning.
When samples are divided into two parts, it may be unnecessary to construct the histograms for both parts. One can store the histogram of their parent node in memory and construct the histogram of one part. Then, the histogram of another part is obtained by subtracting the constructed part from the parent histogram. This trick reduces the running time of histogram construction by at least half.

\section{GBDT for Multiple Outputs}
\label{sec_multi}

In this section, we describe GBDT-MO in detail. We first formulate the general problem of learning GBDT for multiple outputs. Specifically, we derive the objective for learning multiple outputs based on the second order Taylor expansion of loss. We approximate this objective and connect it with the objective for single output. We also formulate the problem of learning a subset of variables and derive its objective. This is achieved by adding $L_0$ constraints. Then, we propose split finding algorithms that minimize the corresponding objectives. Finally, we discuss our implementations and analyze the complexity of our proposed split finding algorithms. In this work, we denote $\mathbf{X}_i$ is $i$-th row of a matrix $\mathbf{X}$, $\mathbf{X}_{.j}$ is $j$-th column and $\mathbf{X}_{ij}$ is its element of $i$-th row and $j$-th column.

\subsection{Objective}
\label{sec_objective_multiple}
We derive the objective of GBDT-MO. Each leaf of a decision tree constructs multiple outputs. Denote $\mathcal{D} = \{ (\mathbf{x}_i, \mathbf{y}_i)_{i=1}^n\}$ as a dataset with $n$ samples, where $\mathbf{x} \in \mathbb{R}^m$ is an $m$ dimensional input and $\mathbf{y} \in \mathbb{R}^d$ is a $d$ dimension output instead of a scalar. Denote $f : \mathbb{R}^m \rightarrow \mathbb{R}^d$ as the function of a decision tree which maps $\mathbf{x}$ into the output space. Based on the construction mechanism of decision trees, $f$ can be further expressed as follows:
\begin{equation}
f(\mathbf{x}) = \mathbf{W}_{q(\mathbf{x})}, \quad q : \mathbb{R}^m \rightarrow [1, L], \quad \mathbf{W} \in \mathbb{R}^{L \times d}
\end{equation}
where $L$ is the number of leaves of a decision tree, $q$ is a function which selects a leaf given $\mathbf{x}$ and $\mathbf{W}_i \in \mathbb{R}^d$ is the values of $i$-th leaf. That is, once a decision tree is constructed, it first maps an input into a leaf, then returns the $d$ dimension vector of that leaf. Next, the prediction of the first $t$ trees is $\hat{\mathbf{y}}_i = \sum_{k=1}^{t} f(\mathbf{x}_i)$.

We consider the objective of the $(t+1)$-th tree given $\hat{\mathbf{y}}$. Because it is separable w.r.t. each leaf (see \eqref{eq_obj_f} and \eqref{eq_obj}), we only consider the objective of a single leaf as follows:
\begin{equation}
\mathcal{L} = \sum_i l(\hat{\mathbf{y}}_i + \mathbf{w}, \mathbf{y}_i) + \lambda \mathcal{R}(\mathbf{w})
\end{equation}
We highlight that $\mathbf{w} \in \mathbb{R}^d$ is a vector with $d$ elements which belongs to a leaf. Again, we suppose $l$ is a second order differentiable function. $l (\hat{\mathbf{y}}_i + \mathbf{w}, \mathbf{y}_i)$ can be approximated by the second order Taylor expansion of $l (\hat{\mathbf{y}}_i, \mathbf{y}_i)$. Set $\mathcal{R}(\mathbf{w}) = \frac{1}{2} \lVert \mathbf{w} \rVert_2^2$, we have:
\begin{equation}
\label{eq_m_obj_taylor}
\mathcal{L} = \sum_i \left\{l (\hat{\mathbf{y}}_i, \mathbf{y}_i) + (\mathbf{g})_i^T\mathbf{w} + \frac{1}{2}\mathbf{w}^T(\mathbf{H})_i\mathbf{w} \right\} + \frac{\lambda}{2}\lVert \mathbf{w} \rVert_2^2
\end{equation}
where $(\mathbf{g})_i = \D{l}{\hat{\mathbf{y}}_i}$ and $(\mathbf{H})_i = \DD{l}{\hat{\mathbf{y}}_i}$. To avoid notation conflicts with the subscript of vectors or matrices, we use $(\cdot)_i$ to indicate that an object belongs to $i$-th sample. This notation is omitted when there is no ambiguity. By setting $\D{\mathcal{L}}{\mathbf{w}} = \mathbf{0}$ for \eqref{eq_m_obj_taylor}, we obtain the optimal leaf values:
\begin{equation}
\label{eq_m_opt_w}
\mathbf{w}^* = -\left( \sum_i (\mathbf{H})_i + \lambda \mathbf{I} \right)^{-1} \left( \sum_i (\mathbf{g})_i \right)
\end{equation}
where $\mathbf{I}$ is an identity matrix. By substituting $\mathbf{w}^*$ into \eqref{eq_m_obj_taylor} and ignoring the constant term $l(\hat{\mathbf{y}}_i, \mathbf{y}_i)$, we get the optimal objective as follows:
\begin{equation}
\label{eq_m_opt_l}
\mathcal{L}^* = -\frac{1}{2} \left( \sum_i (\mathbf{g})_i \right)^T \left( \sum_i (\mathbf{H})_i + \lambda \mathbf{I} \right)^{-1} \left( \sum_i (\mathbf{g})_i \right)
\end{equation}

We have derived the optimal leaf values and the optimal objective for multiple outputs. Comparing \eqref{eq_m_opt_w} with \eqref{eq_opt_w} and \eqref{eq_m_opt_l} with \eqref{eq_opt_l}, it is easy to see that this is a natural generalization of the single output case. In fact, when the loss function $l$ is separable w.r.t. different output dimensions, or equivalently, when its hessian matrix $\mathbf{H}$ is diagonal, each element of $\mathbf{w}^*$ is obtained by the same way as in \eqref{eq_opt_w}.
\begin{equation}
\label{eq_w_app}
\widetilde{\mathbf{w}}^*_j = -\frac{\sum_i (\mathbf{g}_j)_i}{\sum_i (\mathbf{h}_{j})_i + \lambda}
\end{equation}
where $\mathbf{h} \in \mathbb{R}^d$ is the diagonal elements of $\mathbf{H}$. And the optimal objective in \eqref{eq_m_opt_l} can be expressed as the sum of objectives over all output dimensions.
\begin{equation}
\label{eq_diag_opt}
\widetilde{\mathcal{L}}^* = -\frac{1}{2} \sum_{j=1}^{d} \left\{ \frac{\left( \sum_i (\mathbf{g}_j)_i \right)^2}{\sum_i (\mathbf{h}_{j})_i + \lambda} \right\}
\end{equation}
GBDT-MO and GBDT are different even when $\mathbf{H}$ is diagonal because GBDT-MO considers the objectives of all output variables at the same time.

However, it is problematic when $l$ is not separable or equivalently $\mathbf{H}$ is non-diagonal. First, it is difficult to store $\mathbf{H}$ for every sample when the output dimension $d$ is large. Second, to get the optimal objective for each possible split, it is required to compute the inverse of a $d \times d$ matrix, which is time-consuming. Thus, it is impractical to learn GBDT-MO using the exact objective in \eqref{eq_m_opt_l} and the exact leaf values in \eqref{eq_m_opt_w}. Fortunately, it is shown in \cite{chen2015efficient} that $\widetilde{\mathbf{w}}^*$ and $\widetilde{\mathcal{L}}^*$ are good approximations of the exact leaf values and the exact objective when the diagonal elements of $\mathbf{H}$ are dominated. $\widetilde{\mathbf{w}}^*$ and $\widetilde{\mathcal{L}}^*$ are derived from an upper bound of $l(\hat{\mathbf{y}}, \mathbf{y})$. See appendix~\ref{app_proof} for details and further discussions. Although it is possible to derive better approximations for some specific loss functions, we use the above diagonal approximation in this work. We leave better approximations as our future work.

\subsection{Sparse Objective}
In Section \ref{sec_objective_multiple}, we define the objective as the sum of all output variables because there are correlations among variables. However, only a subset of variables are correlated in practice, not all variables. Thus, it is required to learn the values of a suitable subset of $\mathbf{w}$ in a leaf. Although only a subset of variables is covered in a leaf, all variables can be covered because there are many leaves. Moreover, sparse objective is able to reduce the number of parameters. We obtain sparse objective by adding $L_0$ constraint to the non-sparse one. Based on the diagonal approximation, the optimal sparse leaf values are as follows:
\begin{align}
\label{eq_obj_sparse}
\mathbf{w}^*_{sp} = & \arg \min_{\mathbf{w}} G^T\mathbf{w} + \frac{1}{2}\mathbf{w}^T\left(diag(H) + \lambda\mathbf{I} \right) \mathbf{w} \\ \nonumber
s.t. &\quad \lVert \mathbf{w} \rVert_0 <= k
\end{align}
where $k$ is the maximum non-zero elements of $\mathbf{w}$. We denote $G = \sum_i (\mathbf{g})_i$ and $H = \sum_i (\mathbf{h})_i$ to simplify the notation. Note the difference between $H$ and $\mathbf{H}$.

For $j$-th element of $\mathbf{w}_{sp}^*$, its value is either $-\frac{G_j}{H_j + \lambda}$ or 0. Accordingly, the objective contributed by $j$-th column is either $-\frac{1}{2}\frac{G_j^2}{H_j + \lambda}$ or 0. Thus, to minimize \eqref{eq_obj_sparse}, we select $k$ columns with largest $v_j = \frac{G_j^2}{H_j + \lambda}$. Let $\pi$ be the sorted order of $v$ such that:
\begin{equation}
v_{\pi^{-1}(1)} \ge v_{\pi^{-1}(2)} \ge \dots \ge v_{\pi^{-1}(d)}
\end{equation}
Then, the solution of \eqref{eq_obj_sparse} is as follows:
\begin{equation}
(\mathbf{w}_{sp}^*)_j = \left\{
\begin{array}{cc}
-\frac{G_j}{H_j+\lambda}, \qquad & \pi(j) \le k \\
0, \qquad & \pi(j) > k
\end{array}
\right.
\end{equation}
That is, columns with $k$ largest $v$ keep their values while others set to $0$. The corresponding optimal objective is as follows:
\begin{equation}
\label{eq_solution_sparse}
\mathcal{L}^*_{sp} = -\frac{1}{2}\sum_{j: \pi(j) \le k} \frac{G_j^2}{H_j + \lambda}
\end{equation}
Our sparse objective is similar to the objective in \cite{si2017gradient}. Since GBDT-sparse only uses the first order derivative, the second order derivative is set to a constant (1.0) for GBDT-sparse. Thus, $H_j$ in \eqref{eq_solution_sparse} is the number of samples belonging to the leaf. The final solution of the sparse objective only contains a subset of variables. However, to get the optimal subset, all variables are involved in comparisons. That is, all variables are indirectly used.

\begin{algorithm}[!t]
	\caption{Histogram for Multiple Outputs}
	
	\begin{algorithmic}
		\STATE {\bfseries Input:} the set of used samples $\mathcal{S}$, column index $k$, output dimension $d$ and number of bins $b$
		\STATE {\bfseries Output:} histogram of $k$th column
		\STATE Initialize Hist.$\mathbf{g}$ $\in \mathbb{R}^{b\times d}$, Hist.$\mathbf{h}$ $\in \mathbb{R}^{b\times d}$ and Hist.$\mathbf{cnt}$ $\in \mathbb{R}^b$
		\FOR{$i$ {\bfseries in} $\mathcal{S}$}
		\STATE $bin \leftarrow$ bin value of $\mathbf{x}_{ik}$
		\STATE Hist.$\mathbf{cnt}[bin]$ $\leftarrow$ Hist.$\mathbf{cnt}[bin]$ + $1$
		\FOR {$j=1$ {\bfseries to} $d$}
		\STATE Hist.$\mathbf{g}[bin][j]$ $\leftarrow$ Hist.$\mathbf{g}[bin][j]$ + $(\mathbf{g}_j)_i$
		\STATE Hist.$\mathbf{h}[bin][j]$ $\leftarrow$ Hist.$\mathbf{h}[bin][j]$ + $(\mathbf{h}_j)_i$
		\ENDFOR
		\ENDFOR
	\end{algorithmic}
	\label{alg_hist_multi}
\end{algorithm}

\begin{algorithm}[!t]
	\caption{Approximate Split Finding for Multiple Outputs}
	
	\begin{algorithmic}
		\STATE {\bfseries Input:} histograms of current node, input dimension $m$ and output dimension $d$
		\STATE {\bfseries Output:} split with maximum $gain$
		\STATE $gain \leftarrow 0$	
		\FOR {$k=1$ {\bfseries to} $m$}
		\STATE Hist $\leftarrow$ histogram of $k$th column
		\STATE $b$ $\leftarrow$ number of bins of Hist
		\STATE $G \leftarrow \sum_{i=1}^{b}$ Hist.$\mathbf{g}[i]$,
					$H \leftarrow \sum_{i=1}^{b}$ Hist.$\mathbf{h}[i]$
		\STATE $G^l \leftarrow \mathbf{0}$, $H^l \leftarrow \mathbf{0}$
		
		\FOR {$i=1$ {\bfseries to} $b$}
		\STATE $G^l \leftarrow G^l+$ Hist.$\mathbf{g}[i]$,
					$H^l \leftarrow H^l+$ Hist.$\mathbf{h}[i]$
		\STATE $G^r \leftarrow G-G^l$, $H^r \leftarrow H - H^l$
		\STATE $score \leftarrow \sum_{j=1}^{d} \left\{ \frac{(G_j^l)^2}{H_j^l + \lambda} + \frac{(G_j^r)^2}{H_j^r + \lambda} - \frac{(G_j)^2}{H_j+\lambda} \right\}$, \quad \eqref{eq_diag_opt}
		\STATE $gain \leftarrow$ $\max (gain, score)$
		\ENDFOR
		\ENDFOR
		
	\end{algorithmic}
\label{alg_split_appro}
\end{algorithm}

\subsection{Split Finding}
\label{sec_split_multi}
Split finding algorithms for single output maximize the gain of objective before and after split. When dealing with multiple outputs, the objective is defined over all output variables. To find the maximum gain, it is required to scan all columns of outputs. The exact algorithm is inefficient because it enumerates all possible splits. In this work, we use the histogram approximation based one to speed up the training process. To deal with multiple outputs, one should extend the histogram construction algorithm from a single variable case into a multiple variable case. Such an extension is straightforward. Denote $b$ as the number of bins of a histogram. Then, the gradient information is stored in a $b\times d$ matrix and its $j$-th column corresponds to the gradient information of $j$-th variable of outputs. We describe the histogram construction algorithm for multiple outputs in Algorithm \ref{alg_hist_multi}. Once the histogram is constructed, we scan its bins to find the best split. We describe this histogram based split finding algorithm in Algorithm \ref{alg_split_appro}. Compared with the single output one, the objective gain is the sum of gains over all outputs.

\begin{algorithm}[!t]
	\caption{Gain for Sparse Split Finding}
	
	\begin{algorithmic}
		\STATE {\bfseries Input:} gradient statistics of current split, output dimension $d$ and sparse constraint $k$.
		\STATE {\bfseries Output:} gain of current split
		\STATE $Q_l, Q_r \leftarrow$ top-k priority queue
		\FOR {$j=1$ {\bfseries to} $d$}
		\STATE append $\frac{(G^l_j)^2}{H^l_j + \lambda}$ to $Q_l$, append $\frac{(G^r_j)^2}{H^r_j + \lambda}$ to $Q_r$
		\ENDFOR
		\STATE $score = \sum_{v_l \in Q_l} v_l + \sum_{v_r \in Q_r} v_r$, \quad \eqref{eq_loss_s1}
		
	\end{algorithmic}
	\label{alg_gain_two}
\end{algorithm}

\begin{algorithm}[!t]
	\caption{Gain for Restricted Sparse Split Finding}
	
	\begin{algorithmic}
		\STATE {\bfseries Input:} gradient statistics of current split, output dimension $d$ and sparse constraint $k$.
		\STATE {\bfseries Output:} gain of current split
		\STATE $Q \leftarrow$ top-k priority queue
		\FOR {$j=1$ {\bfseries to} $d$}
		\STATE append $\frac{(G^l_j)^2}{H^l_j + \lambda} + \frac{(G^r_j)^2}{H^r_j + \lambda}$ to $Q$
		\ENDFOR
		\STATE $score = \sum_{v \in Q} v$, \quad \eqref{eq_loss_s2}
		
	\end{algorithmic}
	\label{alg_gain_one}
\end{algorithm}

\begin{figure*}
	\centering
	\subfigure[Non-sparse split finding]{
	\includegraphics[width=0.3\textwidth]{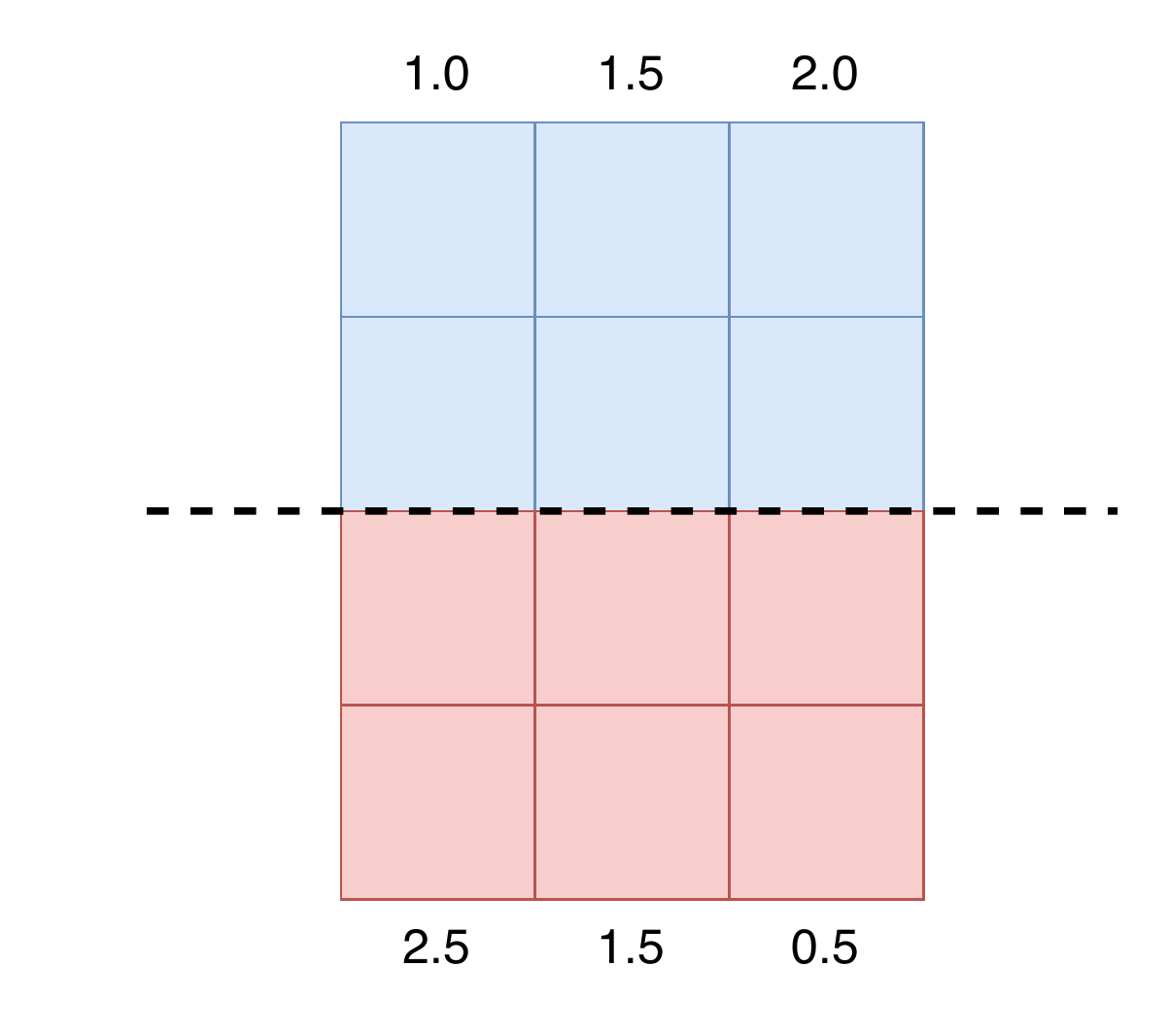}
	}
	\subfigure[Sparse split finding]{
		\includegraphics[width=0.3\textwidth]{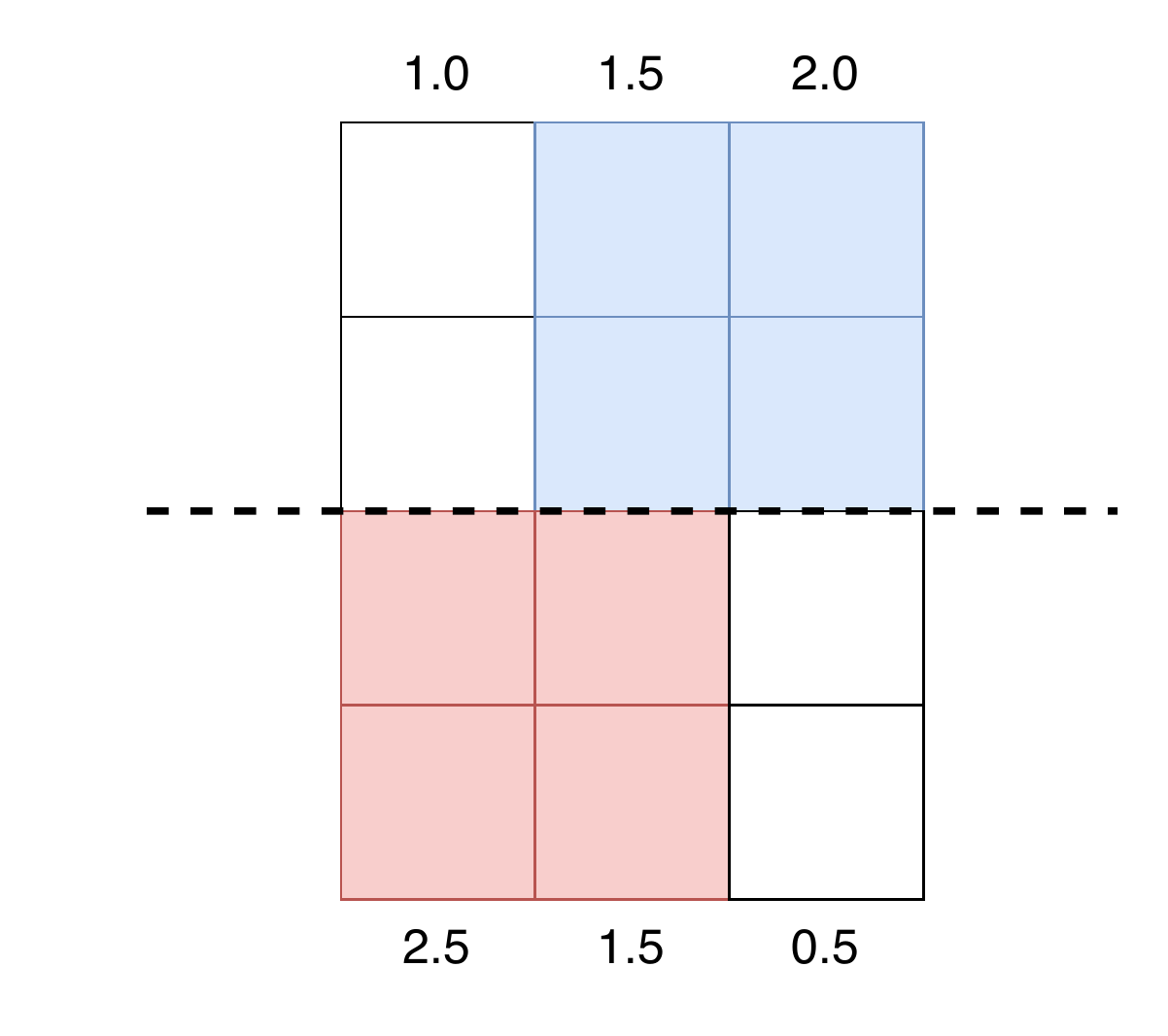}
	}
	\subfigure[Restricted sparse split finding]{
		\includegraphics[width=0.3\textwidth]{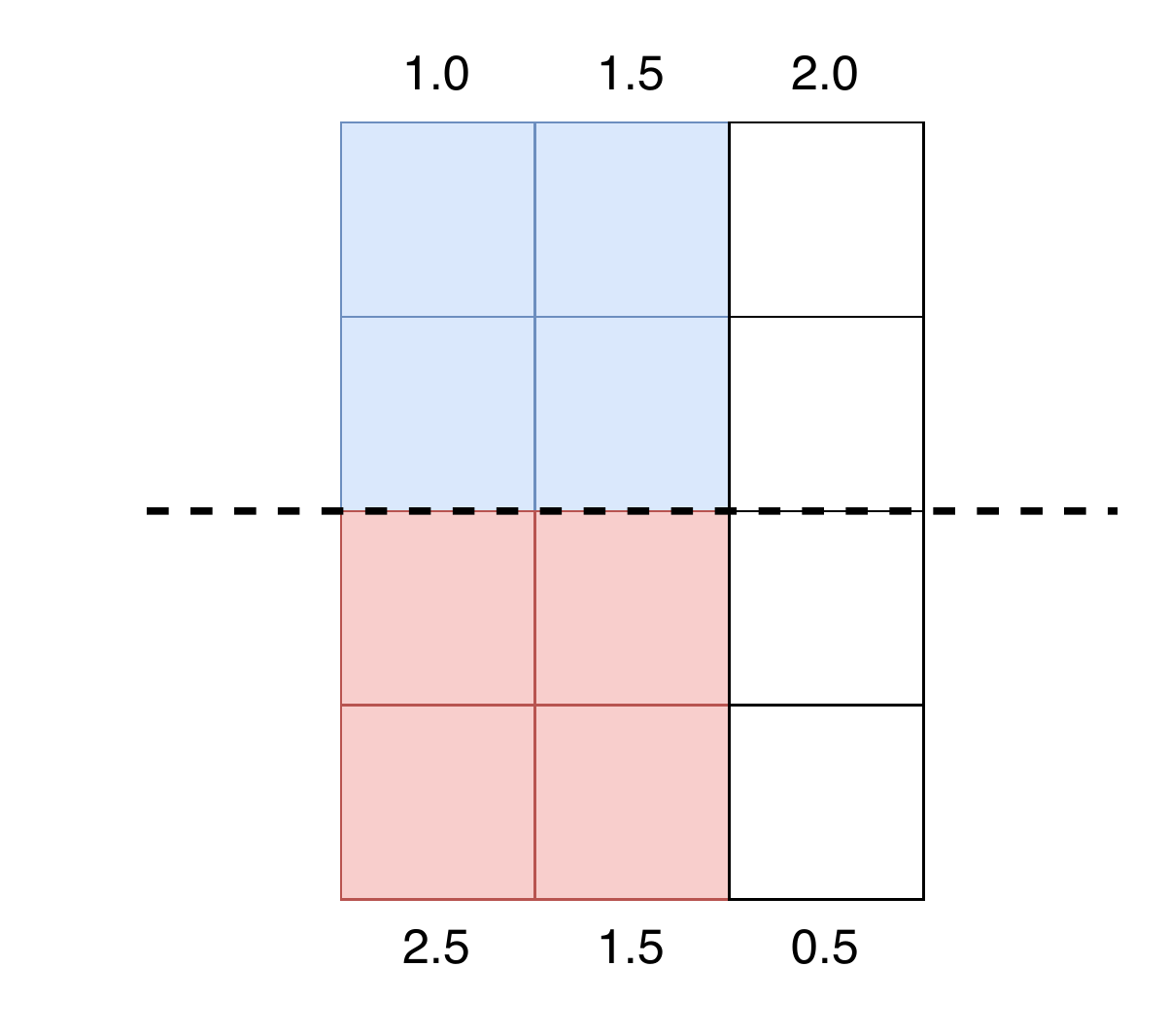}
	}
	\caption{Examples of different split algorithms for $4$ samples in rows and $3$ outputs in columns. The sparse constraint is set to $2$. The dashed line divides samples into two parts. Numbers denote the objectives for each column. Selected columns of left part are marked in blue and selected columns of right part are marked in red. Best viewed on the screen.}
	\label{fig_split}
\end{figure*}

\subsection{Sparse Split Finding}
We propose histogram approximation based sparse split finding algorithms. Compared with the non-sparse one, the key difference is to compute their objective gain given a possible split as follows:
\begin{equation}
\label{eq_loss_s1}
\frac{1}{2} \left\{ \sum_{i: \pi^l(i) \le k} \frac{(G^l_i)^2}{H^l_i + \lambda}  + \sum_{j: \pi^r(j) \le k} \frac{(G^r_j)^2}{H^r_j + \lambda} \right\} - const
\end{equation}
where $\pi^l$ is the sorted order of $\frac{(G^l)^2}{H^l + \lambda}$ and $\pi^r$ is the sorted order of $\frac{(G^r)^2}{H^r + \lambda}$ respectively. $const$ means that the objective before split is fixed for every possible split. When we scan over columns, we maintain the top-k columns for both parts whose $\frac{(G)^2}{H + \lambda}$ is the largest. This can be achieved by a top-k priority queue. We describe the algorithm in Algorithm~\ref{alg_gain_two}.

In Algorithm \ref{alg_gain_two}, the sets of the selected columns of left and right parts are not completely overlapping. We restrict those two sets to be completely overlapping. In other word, the selected columns of two parts are shared. Then, the objective gain in such a case becomes:
\begin{equation}
\label{eq_loss_s2}
\frac{1}{2} \sum_{i: \pi(i) \le k} \left\{ \frac{(G^l_i)^2}{H^l_i + \lambda} + \frac{(G^r_i)^2}{H^r_i + \lambda}\right\} - const
\end{equation}
where $\pi$ is the sorted order of $\frac{(G^l)^2}{H^l + \lambda} + \frac{(G^r)^2}{H^r + \lambda}$. We call it the restricted sparse split finding algorithm. We describe the gain computing for it in Algorithm \ref{alg_gain_one}. There are two advantages of the restricted one:
\begin{itemize}
	\item it has lower computational complexity because it only maintains a single top-k priority queue.
	\item it introduces smoothness prior into the function space because it makes two child nodes with the same parent more similar.
\end{itemize}
We provide an example to show the differences between non-sparse split finding, sparse split finding and restricted sparse split finding in Fig.~\ref{fig_split}.

\begin{algorithm}[!t]
	\caption{Tree Growth}
	
	\begin{algorithmic}
		\STATE {\bfseries Input:} set of samples $\mathcal{S}$, gradient statistics, and hyper-parameters for tree learning.
		\STATE {\bfseries Output:} a decision tree
		\STATE $Hist$ $\leftarrow$ Histograms of $\mathcal{S}$
		\STATE $Split$ $\leftarrow$ $SplitFinding(Hist)$
		\STATE $Q \leftarrow$ priority queue sorted by $Split.gain$
		\STATE $Q.push(Split, Hist, \mathcal{S})$
		\REPEAT
		\STATE $Split, Hist, \mathcal{S} \leftarrow Q.pop()$
		\STATE $Hist^l, Hist^r, \mathcal{S}^l, \mathcal{S}^r \leftarrow ApplySplit(Split, Hist, \mathcal{S})$
		\IF {stop condition is not meet}
		\STATE $Split^l \leftarrow SplitFinding(Hist^l)$
		\STATE $Q.push(Split^l, Hist^l, \mathcal{S}^l)$
		\ENDIF
		\IF {stop condition is not meet}
		\STATE $Split^r \leftarrow SplitFinding(Hist^r)$
		\STATE $Q.push(Split^r, Hist^r, \mathcal{S}^r)$
		\ENDIF
		\UNTIL {$Q$ is empty}

	\end{algorithmic}
	\label{alg_tree}
\end{algorithm}

\subsection{Implementation Details}
We implement GBDT-MO from scratch by C++. First, we implement the core function of LightGBM for ourselves, called GBDT-SO. Then, integrate the learning mechanism for multiple outputs into it, i.e. GBDT-MO\footnote{Our codes are available in \red{https://github.com/zzd1992/GBDTMO.}}. We also provide a Python interface. We speed up GBDT-MO using multi-core parallelism, implemented with OpenMP. A decision tree grows up in the best-first manner. Specifically, we store the information of nodes that has not been divided in memory. At each time when we need to add a node, we select the node whose objective gain is the maximum from all stored nodes. In practice, we store up to $48$ nodes in memory to reduce the memory cost. We describe the algorithm for growth of a tree in Algorithm \ref{alg_tree}.

\subsection{Complexity Analysis}
We analyze the training and inference complexity for GBDT-SO and GBDT-MO. For the training complexity, we focus on the complexity of split finding algorithms. Recall that the input dimension is $m$ and the output dimension is $d$, sparse constraint is $k$, and $b$ is the number of bins. Suppose that $b$ is fixed for all input dimensions, $t$ is the number of boosting rounds and $h$ is the maximum tree depth. Table \ref{tab:complexity_finding} shows the training complexity. The complexity of non-sparse split finding is $\mathcal{O}(bmd)$ because it enumerates on histogram bins, input dimensions and output dimension. For the sparse case, it is multipled by $\log k$ because inserting an element into a top-k priority queue requires $\mathcal{O}(\log k)$ comparisons (see Algorithm \ref{alg_gain_one}). When the exact hessian is used (see \eqref{eq_m_opt_l}), it becomes $\mathcal{O}(bmd^3)$ because calculating the inverse of a $d\times d$ matrix requires $\mathcal{O}(d^3)$ operations. GBDT-SO has the same complexity as GBDT-MO for split finding. However, it does not mean that GBDT-SO is as fast as GBDT-MO. Beyond split finding, GBDT-SO has more overhead that slows its training speed down. For example, samples are divided into two parts, i.e. $d$ times for GBDT-SO, but once for GBDT-MO. Thus, the training speed of GBDT-MO is faster than GBDT-SO in practice. Table \ref{tab:complexity_inference} shows the inference complexity. It can be observed that GBDT-SO needs more comparisons than GBDT-MO.

\begin{table}
	\caption{Complexity of split finding.}\label{tab:complexity_finding}
	\centering
	\begin{tabular}{c|c}
		\hline
		Methods & Complexity\\
		\hline
		\hline
		GBDT-SO & $\mathcal{O}(bmd)$\\
		GBDT-MO & $\mathcal{O}(bmd)$\\
		GBDT-MO (sparse) & $\mathcal{O}(bmd\log k)$\\
		GBDT-MO (exact hessian) & $\mathcal{O}(bmd^3)$\\
		\hline
	\end{tabular}
\end{table}

\begin{table}
	\caption{Complexity of inference.}\label{tab:complexity_inference}
	\centering
	\begin{tabular}{c|c|c}
		\hline
		Methods & Comparisons & Additions\\
		\hline
		\hline
		GBDT-SO  & $\mathcal{O}(tdh)$   & $\mathcal{O}(td)$\\
		GBDT-MO & $\mathcal{O}(th)$ & $\mathcal{O}(td)$\\
		GBDT-MO (sparse) & $\mathcal{O}(th)$ & $\mathcal{O}(tk)$\\
		\hline
	\end{tabular}
\end{table}

\section{Related Work}
\label{sec_related}
Gradient boosted decision tree (GBDT) proposed in \cite{friedman2001greedy} has received much attention due to its accuracy, efficiency and interpretability. GBDT has two characteristics: it uses decision trees as the base learner and its boosting process is guided by the gradient of some loss function. Many variants of \cite{friedman2001greedy} have been proposed. Instead of the first order gradient, XGBoost \cite{chen2016xgboost:} also uses the second order gradient to guide its boost process and derives the corresponding objective for split finding. The histogram approximation of split finding is proposed in \cite{tyree2011parallel} which is used as the base algorithm in LightGBM \cite{ke2017lightgbm:}. Because the second order gradient improves the accuracy and histogram approximation improves the training efficiency, those two improvements are also used in the proposed GBDT-MO.

Since machine learning problems with multiple outputs become common, many tree based or boosting based methods have been proposed to deal with multiple outputs. \cite{agrawal2013multi-label} \cite{prabhu2014fastxml:} generalize the impurity measures defined for binary classification and ranking tasks to a multi-label scenario for splitting a node. However, they are random forest based methods. That is, new trees are not constructed in a boosting manner. Several works extend adaptive boost (AdaBoost) into multi-label cases such as AdaBoost.MH \cite{schapire1998improved} and AdaBoost.LC \cite{amit2007a}. The spirits of AdaBoost.MH and AdaBoost.LC are different from GBDT. At each step, a new base learner is trained from scratch on the re-weighted samples. Moreover, AdaBoost.MH only works for Hamming loss, while AdaBoost.LC only works for the covering loss \cite{amit2007a}.

They do not belong to GBDT families. Two works which belong to GBDT families have been proposed for learning multiple outputs \cite{geurts2007gradient} \cite{si2017gradient}. \cite{geurts2007gradient} transforms the multiple output problem into the single output problem by kernelizing the output space. To achieve this, an $n \times n$ kernel matrix should be constructed where $n$ is the number of training samples. Thus, this method is not scalable. Moreover, it works only for square loss. GBDT for sparse output (GBDT-sparse) is proposed in \cite{si2017gradient}. The outputs are represented in sparse format. A sparse split finding algorithm is designed by adding $L_0$ constraint to the objective. The sparse split finding algorithms of GBDT-MO are inspired by this work. There are several differences between GBDT-MO and GBDT-sparse:
\begin{itemize}
\item GBDT-sparse focuses on extreme multi-label classification problems, whereas GBDT-MO focuses on general multiple output problems. GBDT-sparse requires the loss is separable over output variables. It also requires its gradient is sparse, i.e. $\D{l(\hat{y}, y)}{\hat{y}} = 0$ if $\hat{y} = y$. During training, it introduces the clipping operator into the loss to maintain the sparsity of gradient. The facts limit the types of loss.
\item GBDT-sparse does not employ the second order gradient. Its objective for split finding is derived based on the first order Taylor expansion of the loss as in \cite{friedman2001greedy}.
\item GBDT-sparse does not employ histogram approximation to speed up the training process. It is worthwhile because it is not clear how to construct sparse histograms, especially when combining with the second order gradient.
\item The main motivation of GBDT-sparse is to reduce the space complexity of training and the size of models, whereas the main motivation of GBDT-MO is to improve its generalization ability by capturing correlations between output variables.
\end{itemize}

It may be hard to store the outputs in memory without sparse format when output dimension is very large. In such a situation, GBDT-sparse is a better choice.

\section{Experiments}
\label{sec_exp}
We evaluate GBDT-MO on problems of multi-output regression, multi-class classification and multi-label classification. First, we show the benefits of GBDT-MO using two synthetic problems. Then, we evaluate GBDT-MO on six real-world datasets. We further evaluate our sparse split finding algorithms. Finally, we analyze the impact of diagonal approximation on the performance. Recall that GBDT-SO is our own implementation of GBDT for single output. Except for the split finding algorithm, all implementation details are the same as GBDT-MO for a fair comparison. The purpose of our experiments is not pushing state-of-the-art results on specific datasets. Instead, we would show that GBDT-MO has better generalization ability than GBDT-SO. On synthetic datasets, we compare GBDT-MO with GBDT-SO. On real-world datasets, we compare GBDT-MO with GBDT-SO, XGBoost, LightGBM and GBDT-sparse\footnote{GBDT-sparse is implemented for ourselves, and we do not report its training time due to the slow speed.}. XGBoost and LightGBM are state-of-the-art GBDT implementations for single output, while GBDT-sparse is the most relevant work to GBDT-MO. All experiments are conducted on a workstation with Intel Xeon CPU E5-2698 v4. We use 4 threads on synthetic datasets, while we use 8 threads on real-world datasets. We provide hyper-parameter settings in Appendix~\ref{app_hp}. The training process is terminated when the performance does not improve within 25 rounds.

\begin{table}
	\caption{RMSE on synthetic datasets}
	\label{tab_syn}
	\centering
	\begin{tabular}{c|c|c}
		\hline
		 & friedman1 & random projection\\
		\hline
		\hline
		GBDT-SO & 0.1540 & 0.0204\\
		GBDT-MO & \textbf{0.1429} & \textbf{0.0180}\\
		\hline
	\end{tabular}

\end{table}

\begin{figure*}
	\centering
	\subfigure[friedman1]{
		\includegraphics[width=0.4\textwidth]{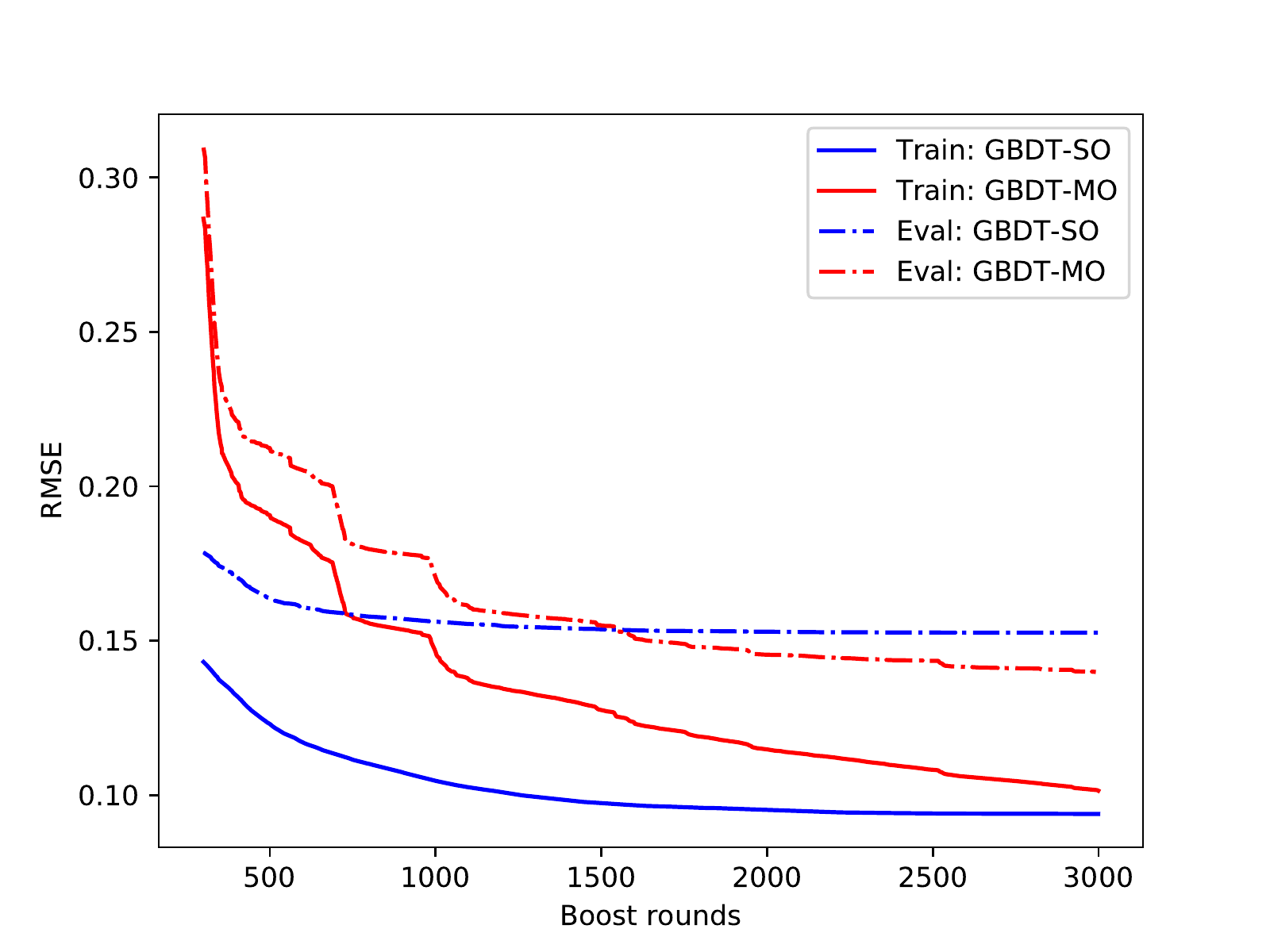}
	}
	\subfigure[random projection]{
		\includegraphics[width=0.4\textwidth]{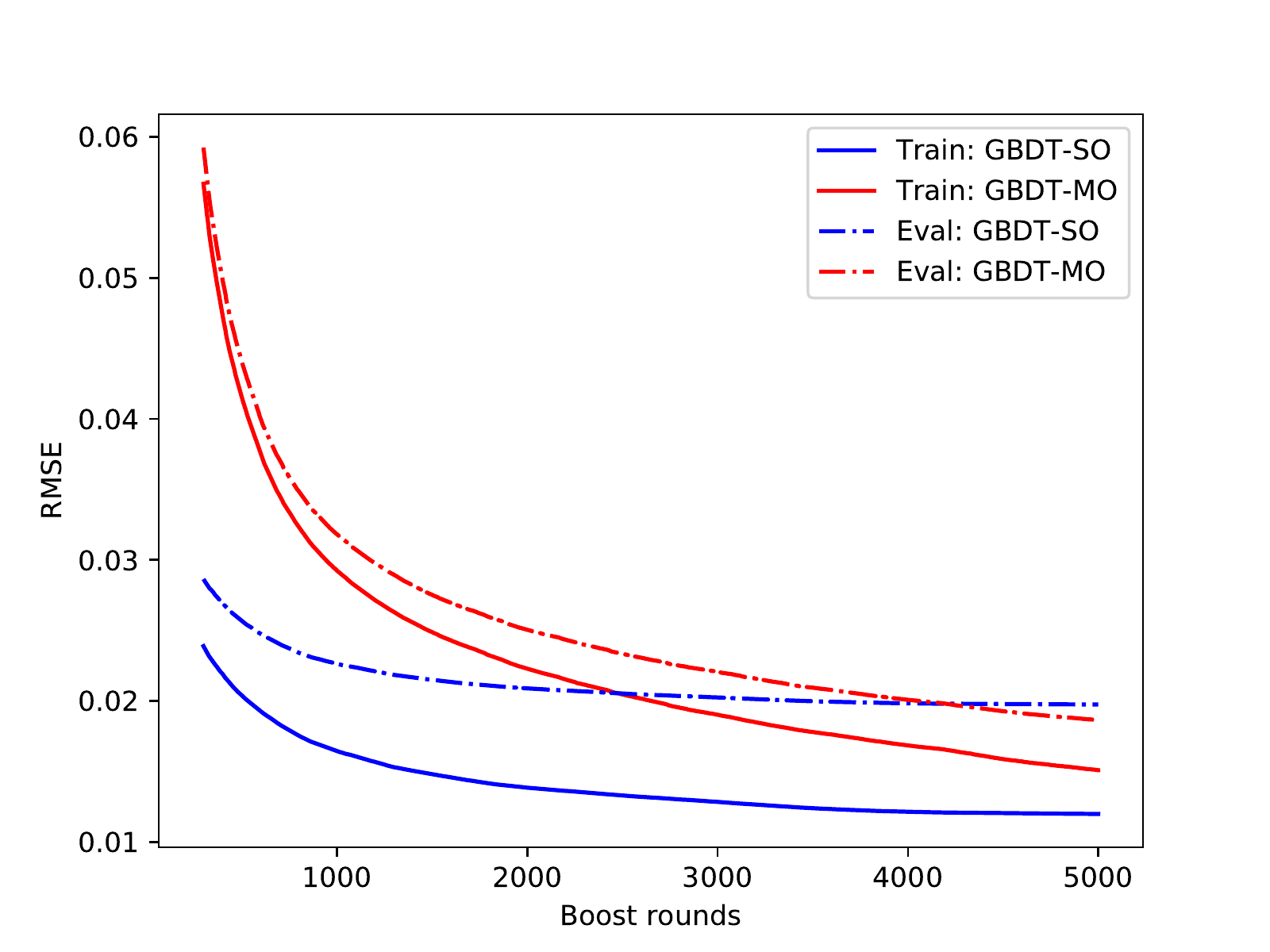}
	}
	\caption{RMSE curves on \textbf{friedman1} and \textbf{random project}. RMSE curves for training samples are drawn in solid lines, while RMSE curves for test samples are drawn in dashed lines. Best viewed on the screen.}
	\label{fig_syn}
\end{figure*}

\begin{figure*}
	\centering
	\subfigure[MNIST]{
		\includegraphics[width=0.4\textwidth]{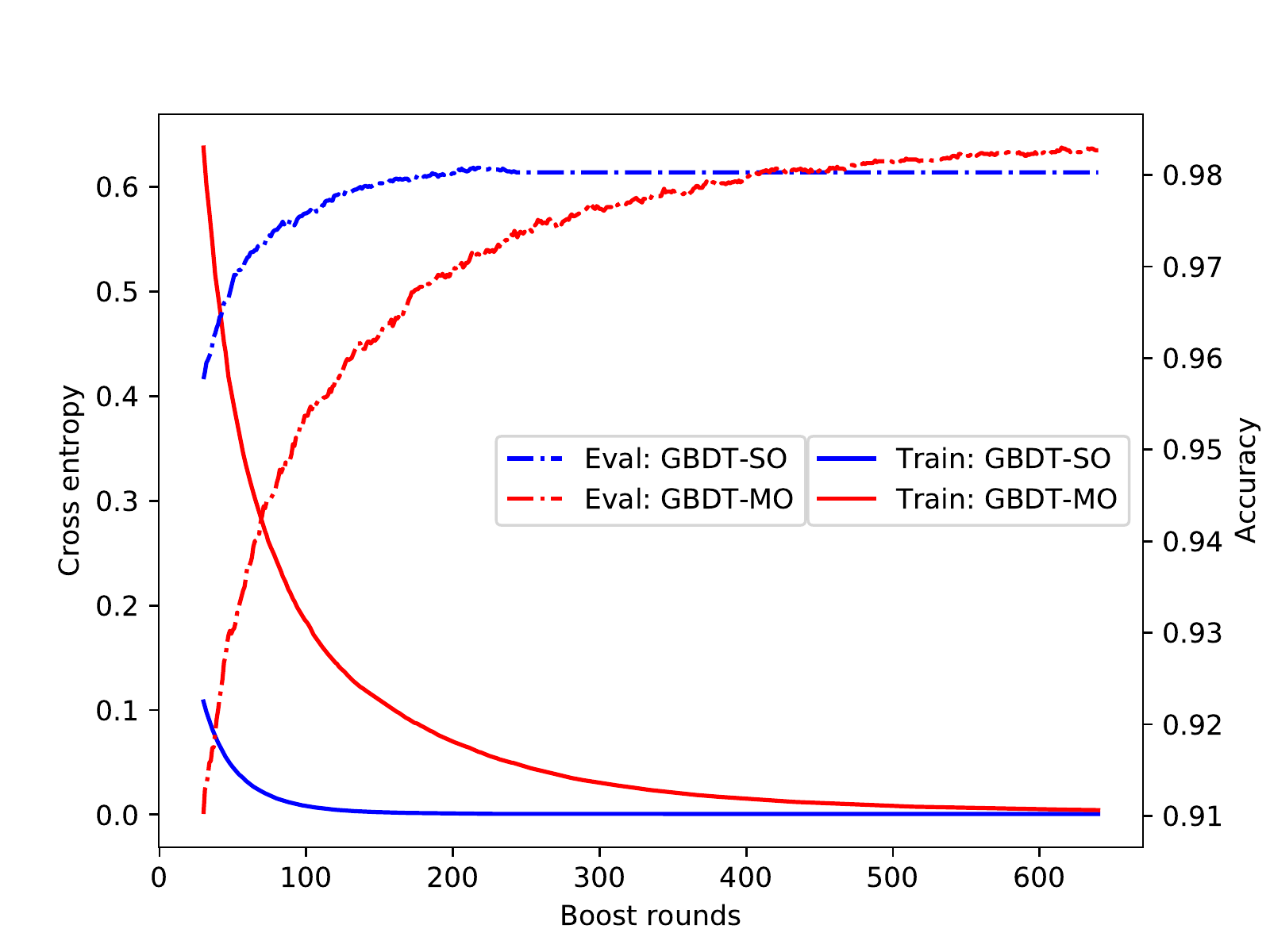}
	}
	\subfigure[Caltech101]{
		\includegraphics[width=0.4\textwidth]{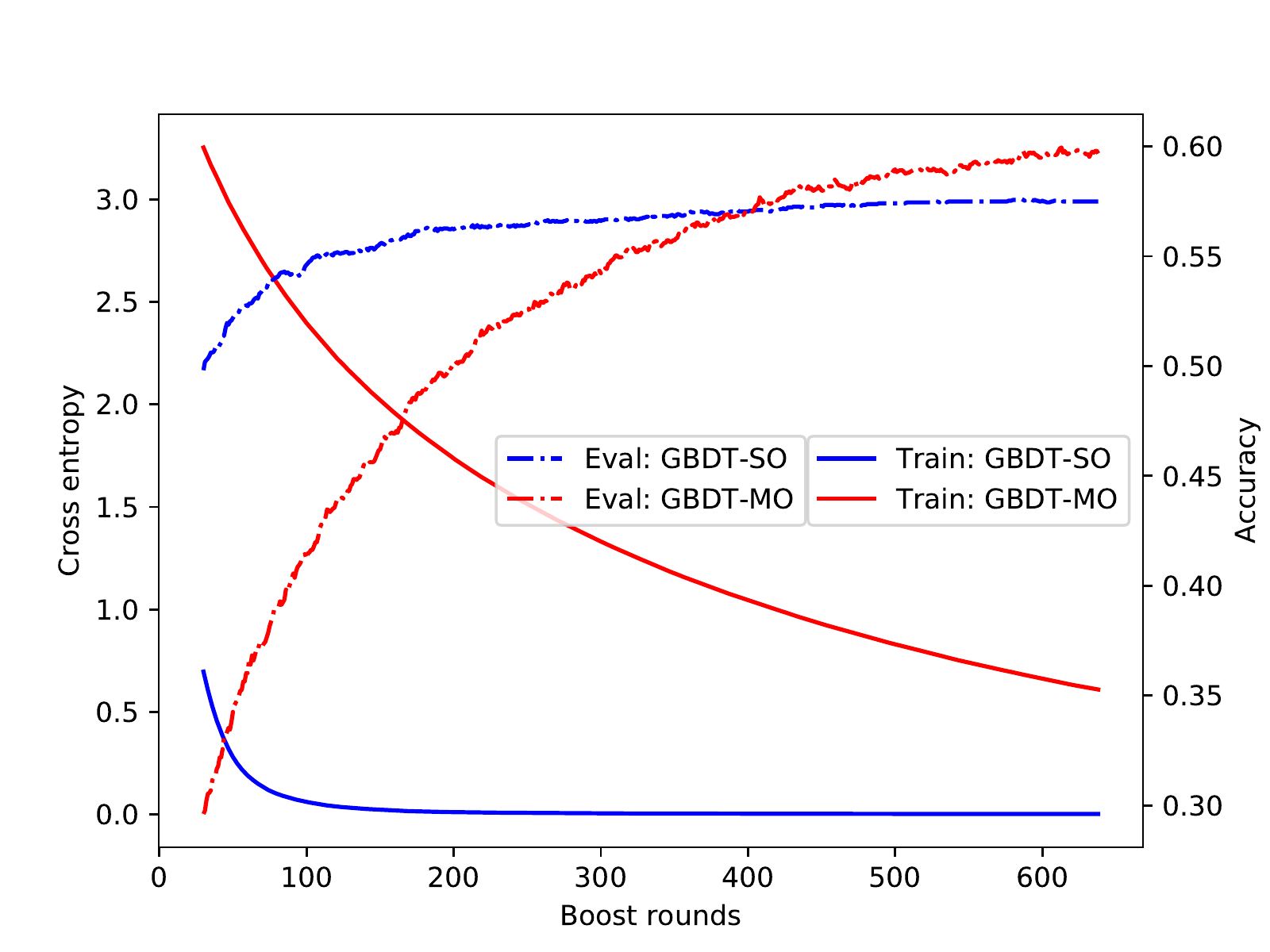}
	}
	\caption{Cross entropy curves and accuracy curves on \textbf{MNIST} and \textbf{Caltech101}. Cross entropy curves for training samples are drawn in solid lines, while accuracy curves for test samples are drawn in dashed lines. Best viewed on the screen.}
	\label{fig_real}
\end{figure*}

\subsection{Synthetic Datasets}
\label{sec_exp_syn}

The first dataset is derived from the \textbf{friedman1} regression problem \cite{friedman1991multivariate}. Its target $y$ is generated by:
\begin{equation}
f(\mathbf{x}) = \sin (\pi \mathbf{x}_1\mathbf{x}_2) + 2(\mathbf{x}_3-0.5)^2 + \mathbf{x}_4 + 0.5\mathbf{x}_5
\end{equation}
\begin{equation}
y = f(\mathbf{x}) + 0.1\varepsilon
\end{equation}
where $\mathbf{x} \in \mathbb{R}^{10}$ and $\varepsilon \sim \mathcal{N}(0; 1)$. Each element of $\mathbf{x}$ is sampled from $\mathcal{U}(-1, 1)$. The last five elements of $\mathbf{x}$ is irrelevant to the target. We extend this problem into multiple output case by adding independent noise to $f(\mathbf{x})$. That is $\mathbf{y}_i = f(\mathbf{x}) + 0.1\varepsilon$ where $\mathbf{y} \in \mathbb{R}^5$.

The second dataset is generated by \textbf{random projection}.
\begin{equation}
\mathbf{y} = \mathbf{w}^T\mathbf{x}
\end{equation}
where $\mathbf{x} \in \mathbb{R}^4$, $\mathbf{w} \in \mathbb{R}^{4 \times 8}$ and $\mathbf{y} \in \mathbb{R}^{8}$. Each element of $\mathbf{x}$ and $\mathbf{w}$ is independently sampled from $\mathcal{U}(-1, 1)$.

For both datasets, we generate $10,000$ samples for training and $10,000$ samples for test. We train them via mean square error (MSE) and evaluate the performance on test samples via root mean square error (RMSE). We repeat the experiments 5 times with different seeds and average the RMSE. As shown in Table \ref{tab_syn}, GBDT-MO is better than GBDT-SO. We provide the training curves in Fig. \ref{fig_syn}. For fairness, curves of different methods are plotted with the same learning rate and maximum tree depth. It can be observed that GBDT-MO has better generalization ability on both datasets. This is because its RMSE on test samples is lower and its performance gap is smaller. Here, performance gap means the differences of performance between training samples and unseen samples which is usually used to measure the generalization ability of machine learning algorithms. The output variables of \textbf{friedman1} are correlated because they are observations of the same underlying variable corrupted by Gaussian noise. The output variables of \textbf{random projection} are also correlated because this projection is over-complete. This supports our claim that GBDT-MO has better generalization abilities because its learning mechanism encourages it to capture variable correlations. However, GBDT-MO suffers from slow convergence speed.

\begin{table*}[!t]
	\caption{Dataset statistics. * means the features are pre-processed.}
	\label{tab_dataset}
	\centering
	\begin{tabular}{c|c|c|c|c|c}
		\hline
		Dataset & \# Training samples & \# Test samples & \# Features & \# Outputs & Problem type\\
		\hline
		\hline
		MNIST & 50,000 & 10,000 & 784 & 10 & classification\\
		Yeast & 1,038 & 446     & 8     & 10 & classification\\
		Caltech101 & 6,073 & 2,604 & 324* & 101 & classification\\
		MNIST-inpainting & 50,000 & 10,000 & 200* & 24 & regression\\		
		Student-por & 454 & 159 & 41* & 3 & regression\\
		NUS-WIDE & 161,789 & 107,859 & 128* & 81 & multi-label\\
		\hline
	\end{tabular}
\end{table*}

\begin{table*}
	\caption{Performance on real-world datasets}
	\label{tab_performance}
	\centering
	\begin{tabular}{c|c|c|c|c|c|c}
		\hline
		\multirow{2}*{} & MNIST & Yeast & Caltech101 & NUS-WIDE & MNIST-inpaining & Student-por\\
		\cline{2-7} & accuracy & accuracy & accuracy & top-1 accuracy & RMSE & RMSE\\
		\hline
		\hline
		XGBoost        & 97.86 &  \textbf{62.94} & 56.52 & 43.72 & 0.26088 & 0.24623\\
		LightGBM       & 98.03 & 61.97  & 55.94 & 43.99 & 0.26090 & 0.24466\\
		GBDT-sparse  & 96.41 & 62.83 & 43.93 & 44.05 & - & -\\
		GBDT-SO       & 98.08 & 61.97  & 56.62 & 44.10 & 0.26157 & 0.24408\\
		GBDT-MO       & \textbf{98.30} & 62.29 & \textbf{57.49} & \textbf{44.21} & \textbf{0.26025} & \textbf{0.24392}\\
		\hline
	\end{tabular}	
\end{table*}

\begin{figure*}[!t]
	\centering
	\includegraphics[width=0.8\textwidth]{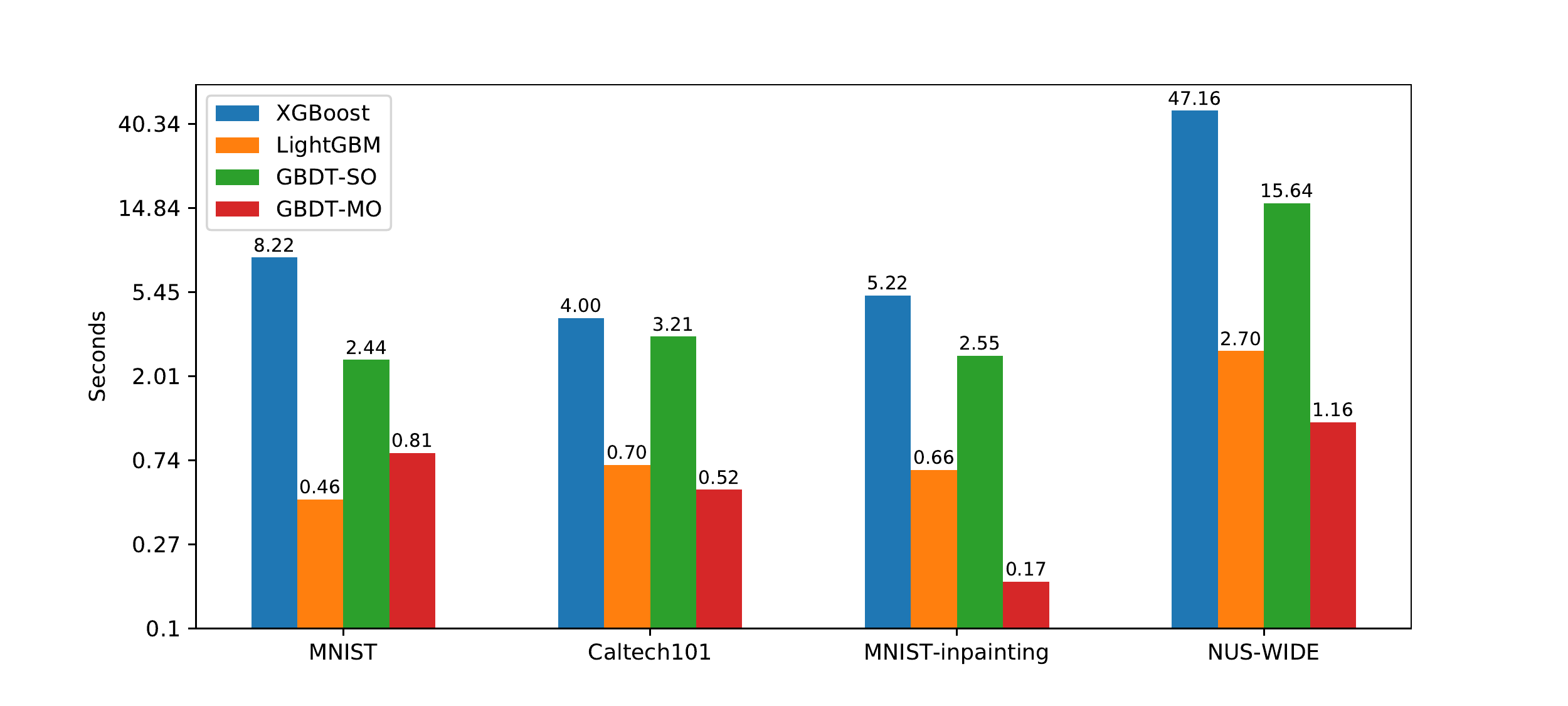}
	\caption{Average training time in second for each boost round. Four relatively larger datasets are used. Y axis is plotted in log scale.}
	\label{fig_time}
\end{figure*}

\begin{table*}[!t]
	\caption{Training time for sparse split finding}
	\label{tab_time_sparse}
	\centering
	\begin{tabular}{c|c|c|c|c}
		\hline
		Restricted/Unrestricted & MNIST & MNIST-inpaining & Caltech101 & NUS-WIDE\\
		\hline
		\hline
		$k=2$    & 0.619/\textbf{0.599} &          -                      &         -            & -\\
		$k=4$    & \textbf{0.737}/0.758 & \textbf{0.149}/0.179  &         -           & -\\
		$k=8$    & \textbf{0.721}/0.778 & \textbf{0.161}/0.195  & \textbf{0.578}/0.663  & \textbf{1.248}/1.266\\
		$k=16$  &      -                          & \textbf{0.153}/0.208 & \textbf{0.620}/0.738  & \textbf{1.237}/1.365\\
		$k=32$  &     -                          &         -                       & \textbf{0.628}/0.866   & \textbf{1.256}/1.453\\
		$k=64$  &     -                          &         -                       & \textbf{0.650}/0.937  & \textbf{1.263}/1.493\\
		\hline
	\end{tabular}	
\end{table*}

\begin{table*}[!t]
	\caption{Performance for sparse split finding}
	\label{tab_performance_sparse}
	\centering
	\begin{tabular}{c|c|c|c|c}
		\hline
		\multirow{2}*{Restricted/Unrestricted} & MNIST & MNIST-inpaining & Caltech101 & NUS-WIDE\\
		\cline{2-5}
		& accuracy & RMSE & accuracy & top-1 accuracy \\
		\hline
		\hline
		$k=2$    & 97.54/\textbf{97.79}  &          -                               &         -                            & -\\
		$k=4$    & \textbf{98.19}/98.18  & 0.26424/\textbf{0.26368}  &         -                            & -\\
		$k=8$    & \textbf{98.07}/98.06 & \textbf{0.26245}/0.26391   & 53.57/\textbf{54.11}  & 44.20/\textbf{44.20}\\
		$k=16$  &      -                               & \textbf{0.26232}/0.26248  & \textbf{55.98}/54.69  & 44.22/\textbf{44.27}\\
		$k=32$  &     -                               &         -                                & \textbf{56.71}/56.64   & \textbf{44.07}/44.04\\
		$k=64$  &     -                               &         -                                & \textbf{58.09}/57.58   & \textbf{44.23}/44.22\\
		\hline
	\end{tabular}	
\end{table*}

\subsection{Real-world Datasets}
\label{sec_exp_real}

In this subsection, we evaluate GBDT-MO on six real-world datasets and compare with related methods in terms of test performance and training speed.

\textbf{MNIST}\footnote{yann.lecun.com/exdb/mnist/} is widely used for classification whose samples are $28\times 28$ gray images of handwritten digits from 0 to 9. Each sample is converted into a vector with 784 elements.

\textbf{Yeast}\footnote{archive.ics.uci.edu/ml/datasets/Yeast} has 8 input attributions each of which is a measurement of the protein sequence. The goal is to predict protein localization sites with 10 possible choices.

\textbf{Caltech101}\footnote{www.vision.caltech.edu/Image\_Datasets/Caltech101/} contains images of objects belonging to 101 categories. Most categories have about 50 samples. The size of each image is roughly $300\times 200$. To obtain fixed length features, we resize each image into $64\times 64$ and compute the HOG descriptor \cite{dalal2005histograms} of the resized image. Each sample is finally converted to a vector with 324 elements.

\textbf{MNIST-inpainting} is a regression task based on \textbf{MNIST}. We crop the central $20\times 20$ patch from the original $28\times 28$ image because most boundary pixels are 0. Then, the cropped image is divided into upper and lower halves each of which is a $10\times 20$ patch. The upper half is used as the input. We further crop a $4 \times 6$ small patch at the top center of the lower half. This small patch is used as the target. The task is to predict the pixels of this $4\times 6$ patch given upper half image.

\textbf{Student-por}\footnote{archive.ics.uci.edu/ml/datasets/Student+Performance} predicts the Portuguese language scores in three different grades of students based on their demographic, social and school related features. The original scores range from 0 to 20. We linearly transform them into $[-1,1]$. We use one-hot coding to deal with the categorical features on this dataset.

\textbf{NUS-WIDE}\footnote{mulan.sourceforge.net/datasets-mlc.html} is a dataset for real-world web image retrieval \cite{nus-wide-civr09}. Tsoumakas et.al \cite{mulan} selects a subset label of this dataset and uses it for multi-label classification. Images are represented using 128-D cVLAD+ features described in \cite{spyromitrosxioufis2014a}.

We summarize the statistics of the above datasets in Table~\ref{tab_dataset}. Those datasets are diverse in terms of scale and complexity. For \textbf{MNIST}, \textbf{MNIST-inpainting} and \textbf{NUS-WIDE}, we use the official training-test split. For others, we randomly select $70\%$ samples for training and the rest for test. We repeat this strategy 10 times with different seeds and report the average results. For multi-class classification, we use cross-entropy loss. For regression and multi-label classification, we use MSE loss. For regression, the performance is measured by RMSE. For multi-class classification and multi-label classification, the performance is measured by top-1 accuracy.

RMSE and accuracy on real-world datasets are shown in Table~\ref{tab_performance}. The overall performance of GBDT-MO is better than others. We provide statistical tests in Appendix~\ref{app_s_test} to show the performance of GBDT-MO on datasets with random training and testing split. We compare the loss curves and the accuracy curves of GBDT-MO and GBDT-SO on \textbf{MNIST} and \textbf{Caltech101} in Fig.~\ref{fig_real}. We conclude that GBDT-MO has better generalization ability than GBDT-SO. Because its training loss is higher while its test performance is better.

We also compare the training speed. Specifically, we run 10 boost rounds and record the average training time for each boost round. We repeat this process three times and report the average training time in seconds as shown in Fig.~\ref{fig_time}. The training speed for \textbf{Yeast} and \textbf{Student-por} is not reported here because those two datasets are too small. GBDT-MO is remarkably faster than GBDT-SO and XGBoost, especially when the number of outputs is large. It is slightly faster than LightGBM. Since we eliminate the interference from outer factors, the comparisons between GBDT-SO and GBDT-MO show the effects of the proposed method objectively.

\subsection{Sparse Split Finding}
All experiments of GBDT-MO performed in Sections \ref{sec_exp_syn} and \ref{sec_exp_real} use non-sparse split finding algorithm. In this subsection, We evaluate our unrestricted sparse split finding algorithm (Algortithm~\ref{alg_gain_two}) and restricted sparse split finding algorithm (Algorithm~\ref{alg_gain_one}). We compare their performance and training speed on \textbf{MNIST}, \textbf{MNIST-inpainting}, \textbf{Caltech101} and \textbf{NUS-WIDE} with different sparse factor $k$. The hyper-parameters used here are the same as the corresponding non-sparse one's.

Their training speed in seconds is shown in Table~\ref{tab_time_sparse}. The restricted one is faster than the unrestricted one. Their speed differences are remarkable when $k$ is large. The performance on test samples is shown in Table~\ref{tab_performance_sparse}. The restricted one is slightly better than the unrestricted one. Note that the performance of our sparse split algorithms is sometimes better than the non-sparse one's with a proper $k$ (for example, $k=64$ on \textbf{Caltech101}).

\subsection{Objective with Exact Hessian}
GBDT-MO replaces the exact hessian with the diagonal hessian to reduce computational complexity. We analyze the impact of the diagonal approximation on \textbf{MNIST} and \textbf{Yeast} because their output dimensions are relatively small. Table~\ref{tab_exactH} shows accuracy and training time for a single round. Two methods have similar accuracy but the diagonal approximation is much faster. Thus, the diagonal approximation of hessian achieves good trade-off between accuracy and speed.

\begin{table}
	\caption{Impact of diagonal hessian.}
	\label{tab_exactH}
	\centering
	\begin{tabular}{c|c|c}
		\hline
		Accuracy/Time & MNIST & Yeast\\
		\hline
		\hline
		diagonal hessian  &  98.30/0.79s  & 62.29/0.0011s\\
		exact hessian       &  98.17/3.96s  & 62.89/0.0094s\\
		\hline
	\end{tabular}
	
\end{table}

\section{Conclusions}
\label{sec_conc}
In this paper, we have proposed a general method to learn GBDT for multiple outputs. The motivation of GBDT-MO is to capture the correlations between output variables and reduce redundancy of tree structures. We have derived the approximated learning objective for both non-sparse case and sparse case based on the second order Taylor expansion of loss. For sparse case, we have proposed the restricted algorithm which restricts the subsets of left and right parts to be the same and the unrestricted algorithm which has no such a restriction. We have extended the histogram approximation into multiple output case to speed up the training process. We have evaluated that GBDT-MO is remarkably and consistently better in generalization ability and faster in training speed compared with GBDT for single output. We have also evaluated that the restricted sparse split finding algorithm is slightly better than the unrestricted one by considering both performance and training speed. However, GBDT-MO suffers from slower convergence, especially at the beginning of training. In our future work, we would improve its convergence speed.

\ifCLASSOPTIONcaptionsoff
  \newpage
\fi

\bibliographystyle{IEEEbib}
\bibliography{reference}




\newpage
\appendices

\begin{table*}[t]
	\caption{Grid searched learning rate and maximum depth}
	\label{tab_d_lr}
	\centering
	\begin{tabular}{c|c|c|c|c|c|c}
		\hline
		Learning rate/Maximum depth & MNIST & Yeast & Caltech101 & NUS-WIDE & MNIST-inpaining & Student-por\\
		\hline
		\hline
		XGBoost            & 0.10/8 & 0.10/4 & 0.10/8 & 0.10/8 & 0.10/8 & 0.10/4\\
		LightGBM          & 0.25/6 & 0.10/4 & 0.10/9 & 0.05/8 & 0.10/8 & 0.05/4\\
		GBDT-sparse     & 0.05/8 & 0.10/5 & 0.10/9 & 0.10/8 & - & -\\
		GBDT-SO           & 0.10/6 & 0.10/4 & 0.05/8 & 0.05/8 & 0.10/8 & 0.05/4\\
		GBDT-MO          & 0.10/8 & 0.10/5 & 0.10/10 & 0.10/8 & 0.10/7 & 0.10/4\\
		\hline
	\end{tabular}	
\end{table*}

\begin{table*}[t]
	\caption{Number of trees of learned models}
	\label{tab_num_tree}
	\centering
	\begin{tabular}{c|c|c|c|c|c|c}
		\hline
		& MNIST & Yeast & Caltech101 & NUS-WIDE & MNIST-inpaining & Student-por\\
		\hline
		\hline
		XGBoost            & 1700    & 299.0   & 2234.0  & 5921     & 4880   & 133.4\\
		LightGBM           & 1760   & 334.0    & 2692.0  & 11636   & 5618   & 219.6\\
		GBDT-sparse     & 645     & 21.4      & 237.4    &  494      & -         & -\\
		GBDT-SO           & 2180   & 183.0    & 3697.0  &  12051   & 6373   & 373.7\\
		GBDT-MO          & 615     & 85.1      & 517.7     &  3546    & 2576   & 125.0\\
		\hline
	\end{tabular}	
\end{table*}

\begin{table*}[t]
	\caption{Preliminarily searched hyper-parameters}
	\label{tab_hp}
	\centering
	\begin{tabular}{c|c|c|c|c|c|c}
		\hline
		& MNIST & Yeast & Caltech101 & NUS-WIDE & MNIST-inpaining & Student-por\\
		\hline
		\hline
		Gain threshold          & 1e-3 & 1e-3 & 1e-3 & 1e-6 & 1e-6 & 1e-6\\
		Min samples in leaf   & 16 & 16 & 16 & 4 & 4 & 4\\
		Max bins                  & 8 & 32 & 32 & 64 & 16 & 8\\
		\hline
	\end{tabular}	
\end{table*}

\section{Approximated Objective}
\label{app_proof}
We suppose there exists $\gamma > 0$ such that
\begin{eqnarray}
\label{eq_assum}
\gamma \mathbf{H}_{ii} \ge \sum_{j} |\mathbf{H}_{ij}|, \quad \forall i
\end{eqnarray}
That is, $\mathbf{H}$ is dominated by its diagonal elements. Then, we have:
\begin{align}
\mathbf{w}^T\mathbf{Hw} & = \sum_{i} \sum_{j} \mathbf{H}_{ij}\mathbf{w}_i\mathbf{w}_j \\ \nonumber
& \le \frac{1}{2}\sum_{i}\sum_{j} | \mathbf{H}_{ij} | (\mathbf{w}_i^2 + \mathbf{w}_j^2) \\ \nonumber
& = \sum_{i}\sum_{j} | \mathbf{H}_{ij} | \mathbf{w}_i^2 \\ \nonumber
& \le \gamma \sum_i \mathbf{H}_{ii} \mathbf{w}_i^2
\end{align}
The last inequality holds based on the assumption in \eqref{eq_assum}. Substituting this result into the Taylor expansion of loss
\begin{align}
l(\mathbf{\hat{y}} + \mathbf{w}, \mathbf{y}) &\approx l(\mathbf{\hat{y}}, \mathbf{y}) + \mathbf{g}^T\mathbf{w} + \frac{1}{2}\mathbf{w}^T\mathbf{Hw} \\ \nonumber
& \le l(\mathbf{\hat{y}}, \mathbf{y}) + \sum_{i} \mathbf{g}_i\mathbf{w}_i + \frac{\gamma}{2}\sum_i \mathbf{H}_{ii}\mathbf{w}_i^2
\end{align}
We ignore the remainder term of the Taylor expansion~\cite{chen2016xgboost:}. Substituting this upper bound of $l(\mathbf{\hat{y}} + \mathbf{w}, \mathbf{y})$ into \eqref{eq_m_obj_taylor}, we get the optimal value of $\mathbf{w}$.
\begin{equation}
\mathbf{w}^*_j = -\frac{\sum_i (\mathbf{g}_j)_i}{\gamma \sum_i (\mathbf{h}_{j})_i + \lambda}
\end{equation}
Recall that $\mathbf{h}$ is the diagonal elements of $\mathbf{H}$. This solution is the same as \eqref{eq_w_app}, except the coefficient $\gamma$. In practice, the actual leaf weight is $\alpha\mathbf{w}^*_j$ where $\alpha$ is the so-called learning rate. Then, the effect of $\gamma$ can be canceled out by adjusting $\alpha$ and $\lambda$. Thus, we do not need to consider the exact value of $\gamma$ in practice.

\section{Hyper-parameter Settings}
\label{app_hp}

We discuss our hyper-parameter settings. We find the best maximum depth $d$ and learning rate via a grid search. $d$ is searched from $\{4, 5, 6, 7, 8, 9, 10\}$ and learning rate is searched from $\{0.05, 0.1, 0.25, 0.5\}$. We provide the selected $d$ and learning rate in Table~\ref{tab_d_lr}. We set $L_2$ regularization $\lambda$ to 1.0 and the maximum leaves to $0.75 \times 2^d$ in all experiments. We preliminarily search other hyper-parameters and fix them for all methods. We list them on real-world datasets in Table~\ref{tab_hp}. Note that the hyper-parameters in Table~\ref{tab_hp} may not be optimal. When comparing with the gain threshold, we use the average gain over the involved output variables. We provide the number of trees of the best models in Table \ref{tab_num_tree}. For single variable GBDT models, the number of trees is the summation of the number of trees for each output variable.

\section{Statistical Test of Results}
\label{app_s_test}
There are no standard training and testing splits on \textbf{Yeast}, \textbf{Caltech101} and \textbf{Student-por}. We obtain the results in Table \ref{tab_performance} by averaging the results of 10 random trials. Thus, statistical tests on how likely GBDT-MO is better than others are necessary. Denote $X$ is a random variable which means the performance differences between GBDT-MO and the others. $X$ is commonly supposed to be Student's T-distribution because its true variance is unknown. We estimate the mean and standard deviation of $X$ from the 10 random trials. Then, the confidence in whether GBDT-MO performs better than the others is defined as $P(X>0)$ for classifications and $P(X<0)$ for regressions, as reported in Table \ref{tab_T_test}. Most confidence scores are significantly higher than 0.5. Thus, it can be expected that GBDT-MO performs better than others.

\begin{table}[t]
	\caption{Confidence of the superiority of GBDT-MO}
	\label{tab_T_test}
	\centering
	\begin{tabular}{c|c|c|c}
		\hline
		Confidence  & Yeast & Caltech101 & Student-por\\
		\hline
		\hline
		XGBoost      & 0.067 & 0.930 & 0.888\\
		LightGBM     & 0.745 & 0.988 & 0.685\\
		GBDT-SO     & 0.812 & 0.955 & 0.561\\
		\hline
	\end{tabular}	
\end{table}

\end{document}